\documentclass[10pt,twocolumn,letterpaper]{article}

\usepackage[pagenumbers]{cvpr} 

\usepackage{amssymb}
\DeclareMathOperator*{\argmax}{arg\,max}

\definecolor{cvprblue}{rgb}{0.21,0.49,0.74}
\usepackage[table]{xcolor}
\usepackage{graphicx}
\usepackage{multirow}
\usepackage{mathtools}
\usepackage{marvosym}

\usepackage[pagebackref,breaklinks,colorlinks,allcolors=cvprblue]{hyperref}

\def\paperID{5218} 
\def\confName{CVPR}
\def\confYear{2026}

\title{Robo-Dopamine: General Process Reward Modeling for High-Precision \\ Robotic Manipulation}

\author{\vspace{0.25em} Huajie Tan$^{1,2,*,\dagger}$, Sixiang Chen$^{1,2,*}$, Yijie Xu$^{2,3,*}$, Zixiao Wang$^{1,2}$,  Yuheng Ji$^{2,4}$, \\
\vspace{0.25em} Cheng Chi$^{2}$, Yaoxu Lyu$^{1,2}$, Zhongxia Zhao$^{1,2}$, Xiansheng Chen$^{2}$, Peterson Co$^{1,2}$, \\
Shaoxuan Xie$^{2}$, Guocai Yao$^{2}$, Pengwei Wang$^{2,\dagger}$, Zhongyuan Wang$^2$, Shanghang Zhang$^{1,2,\text{\Letter}}$ \\
$^1$ \small State Key Laboratory of Multimedia Information Processing, School of Computer Science, Peking University \\ 
\vspace{0.2em}$^2$ \small Beijing Academy of Artificial Intelligence,
$^3$ \small University of Sydney, 
$^4$ \small Institute of Automation, Chinese Academy of Sciences \\
\textit{\textbf{Project website:}} \href{https://robo-dopamine.github.io/}{https://robo-dopamine.github.io}
}

\begin{document}

\twocolumn[{%
\maketitle
\vspace{-0.9cm}
\begin{center}
    \centering
    \captionsetup{type=figure}
    \includegraphics[width=0.99\linewidth]{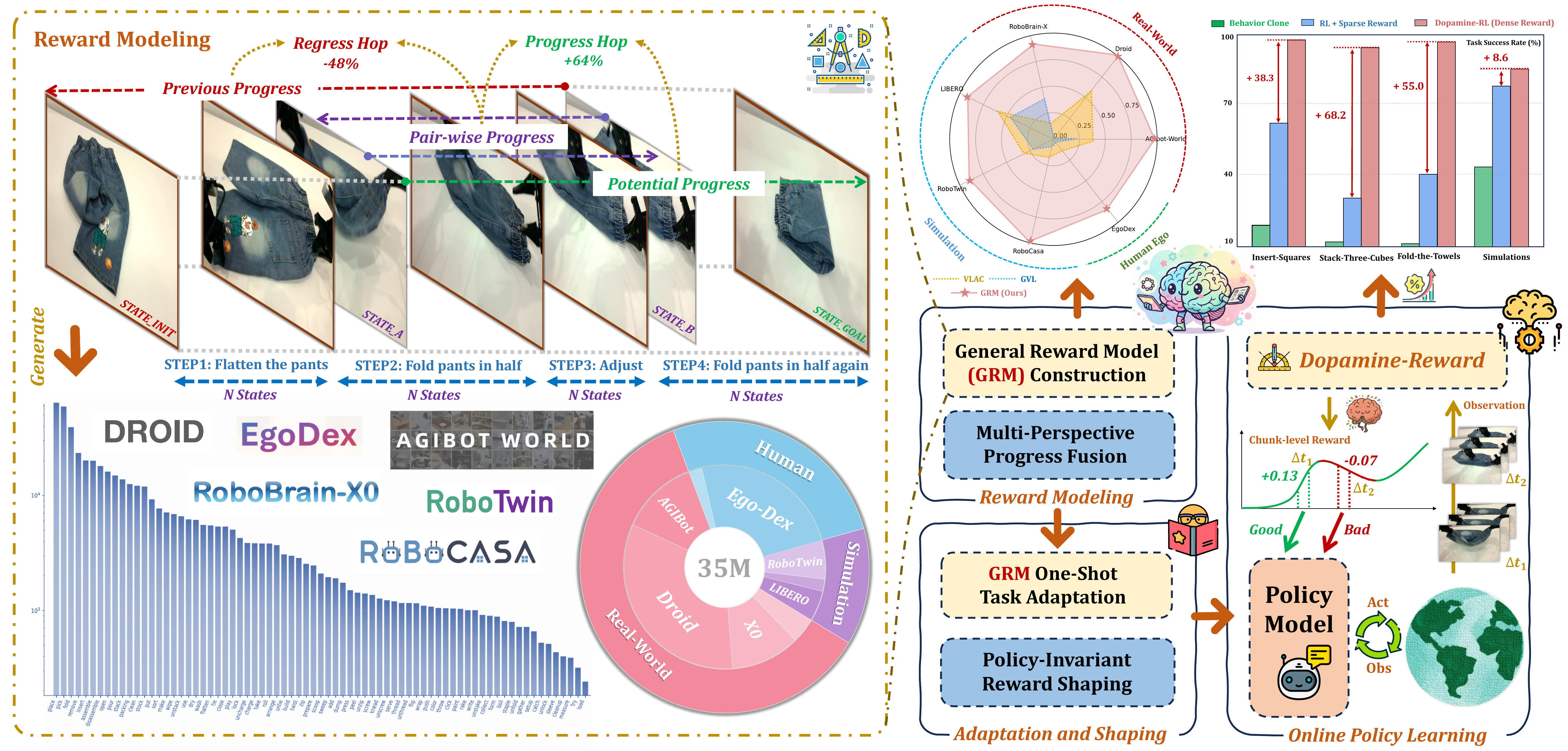}
    \captionof{figure}{
    \textbf{Overview of Robo-Dopamine.} Robo-Dopamine integrates large-scale reward modeling with a robust policy learning algorithm.
    \textbf{\textit{(Left)}} We construct a General Reward Model (GRM) trained on a large and diverse 35M-sample dataset spanning real-world, simulation, and human-centric videos with our Dopamine-Reward, a step-aware fine-grained reward modeling method. This GRM learns to predict fine-grained, relative progress between states to accurately assess task progression.
    \textbf{\textit{(Bottom Right)}} The pre-trained GRM is adapted to new tasks and provides dense reward signals to our Dopamine-RL framework. By using a theoretically-sound Policy-Invariant Reward Shaping method, Dopamine-RL efficiently guides the policy during online interactions without misaligning the task objective.
    \textbf{\textit{(Top Right)}} Our integrated approach establishes a new state-of-the-art in reward accuracy (radar chart) and demonstrates high training efficiency, significantly boosting policy success rates in both simulation and the real world (bar chart).
    } 
    \label{fig:teaser}
\end{center}
}]

\let\thefootnote\relax\footnotetext{$^{*}$ Equal contribution.}
\let\thefootnote\relax\footnotetext{$^{\dagger}$ Project leaders.}
\let\thefootnote\relax\footnotetext{$^{\text{\Letter}}$ Corresponding author: \href{shanghang@pku.edu.cn}{shanghang@pku.edu.cn}}

\maketitle
\begin{abstract}
The primary obstacle for applying reinforcement learning (RL) to real-world robotics is the design of effective reward functions. While recently learning-based Process Reward Models (PRMs) are a promising direction, they are often hindered by two fundamental limitations: 
their reward models lack step-aware understanding and rely on single-view perception, leading to unreliable assessments of fine-grained manipulation progress; and their reward shaping procedures are theoretically unsound, often inducing a semantic trap that misguides policy optimization.
To address these, we introduce Dopamine-Reward, a novel reward modeling method for learning a general-purpose, step-aware process reward model from multi-view inputs. At its core is our General Reward Model (GRM), trained on a vast 3,400+ hour dataset, which leverages Step-wise Reward Discretization for structural understanding and Multi-Perspective Reward Fusion to overcome perceptual limitations. Building upon Dopamine-Reward, we propose Dopamine-RL, a robust policy learning framework that employs a theoretically-sound Policy-Invariant Reward Shaping method, which enables the agent to leverage dense rewards for efficient self-improvement without altering the optimal policy, thereby fundamentally avoiding the semantic trap.
Extensive experiments across diverse simulated and real-world tasks validate our approach. {GRM achieves state-of-the-art accuracy in reward assessment}, and {Dopamine-RL built on GRM significantly improves policy learning efficiency}.
For instance, after {GRM} is adapted to a new task in a one-shot manner from a single expert trajectory, the resulting reward model enables Dopamine-RL to improve the policy from near-zero to 95\% success with only 150 online rollouts (approximately 1 hour of real robot interaction), while retaining strong generalization across tasks. Project website: \href{https://robo-dopamine.github.io/}{Robo-Dopamine}.
\end{abstract}
    
\section{Introduction}
\label{sec:intro}

While large-scale imitation learning (IL) has substantially advanced embodied intelligence \cite{abdolmaleki2025gemini, intelligence2025pi_, Helix2024, kim2025oft, bjorck2025gr00t, bai2025towards}, its reliance on static, expert-curated datasets imposes fundamental limitations \cite{generalist2025gen0, khazatsky2024droid, zhou2025roborefer, zhou2025robotracer, o2024open, bu2025agibot}, which exhibits sub-optimal sample efficiency, poor generalization to out-of-distribution (OOD) scenarios, and also struggles to acquire precise and contact-rich manipulation skills \cite{jiang2025dexmimicgen, zheng2025x}. In contrast, reinforcement learning (RL) offers a compelling alternative \cite{chen2025conrft, luo2025precise, lu2025vla, zhang2025reinbot, xu2024rldg, li2025simplevla, zang2025rlinf}. Through continuous environmental interaction, RL enables agents to transcend the limitations of static expert data, facilitating superior generalization and the mastery of high-precision tasks.

However, the primary obstacle for applying RL to real-world robotics is the design of effective reward functions. Conventional approaches falter at two extremes: sparse, binary outcome rewards \cite{chen2025conrft, luo2025precise, lu2025vla, zhang2025reinbot, xu2024rldg} make exploration in long-horizon, contact-rich tasks prohibitively difficult, while handcrafted dense rewards \cite{peng2022deep, wu2021learning, finn2016guided, van2024revisiting} require significant domain expertise, limiting scalability and general applicability. This dichotomy has motivated the shift towards learning-based Process Reward Models (PRMs) \cite{ma2023liv, alakuijala2024video, ma2024gvl, zhai2025vlac, chen2025sarm}. Despite their promise, current PRMs are hindered by two fundamental limitations. \textit{First,} the underlying reward models often exhibit critical deficiencies: their task-specific design \cite{ghasemipour2025self, chen2025sarm} inherently limits generalization; uniform reward distributions \cite{ma2024gvl, zhai2025vlac} fail to capture the varying salience of crucial sub-steps; 
and a reliance on single-view observations \cite{ma2023liv, alakuijala2024video, ma2024gvl, zhai2025vlac, chen2025sarm} fails in manipulation scenes where occlusions obscure fine-grained progress only visible from wrist-level views.
\textit{Second,} the reward shaping algorithms utilizing these dense signals are often theoretically flawed. Naively incorporating dense rewards can induce a semantic trap \cite{ng1999policy} that misguides policy optimization by inadvertently altering the optimal policy, causing the agent to prioritize high proxy rewards from intermediate steps over the true task objective.

To address these, we introduce \textbf{Dopamine-Reward}, a novel dense reward modeling method for learning a general-purpose, step-aware process reward from multi-view inputs. Dopamine-Reward directly tackles the first limitation by leveraging two key techniques: Hop-based Step-wise General Reward Model (GRM) Construction for a fine-grained, structural understanding of task progression from various viewpoints, and Multi-Perspective Reward Fusion via GRM to integrate bidirectional global reward and state-wise incremental reward for more precise reward estimation, which are made possible by a meticulous annotation pipeline encompassing over 3,400 hours of data, 100K trajectories, and more than 350 daily tasks, offering broad coverage, fine-grained labels, and well-balanced distributions across real robots, simulations, and egocentric human videos.

Building upon GRM via Dopamine-Reward, we propose a robust and unified policy learning framework \textbf{Dopamine-RL} to resolve the second limitation. Dopamine-RL employs a theoretically-sound Policy-Invariant Reward Shaping method, which enables the agent to leverage the dense rewards from our GRM for highly efficient self-improvement without altering the underlying optimal policy, thereby fundamentally avoiding the semantic trap. Extensive experiments on over 10 simulation and 8 real-world tasks demonstrate the superiority of our methods:
(1) \textit{\textbf{State-of-the-art Reward Accuracy.}} The GRM achieves over 92.8\% accuracy in progress assessment, with a Value-Order Consistency (VOC) score of 0.953 on rank-correlation benchmarks, outperforming established baselines.
(2) \textit{\textbf{High Training Efficiency.}}
After GRM is adapted to a new task in a one-shot manner from a single expert demonstration, the resulting reward model enables Dopamine-RL to improve a policy from near-zero to 95\% success rate within approximately 150 online rollouts (about one hour of real robot interaction), with some tasks reaching 100\% success rate.
(3) \textit{\textbf{Improved Generalization.}} 
By combining step-wise structural modeling, reward fusion, and multi-view perception for robust estimation under occlusion and fine-grained state changes, our GRM provides more reliable learning signals, enabling Dopamine-RL to generalize more effectively to unseen layouts, backgrounds, and object variations.

An overview of Dopamine-Reward and Dopamine-RL, together with our empirical gains in reward accuracy and policy performance, is shown in Figure \ref{fig:teaser}. In summary, our main contributions are as follows:
\begin{itemize}

    \item We propose \textit{Dopamine-Reward}, a novel reward modeling method built around a {General Reward Model (GRM)} that provides step-aware, fine-grained, and occlusion-resilient process rewards for precise robotic manipulation.

    \item We introduce \textit{Dopamine-RL}, a robust policy learning framework with a theoretically grounded Policy-Invariant Reward Shaping scheme, which effectively exploits dense GRM rewards to accelerate policy optimization while avoiding the semantic trap.

    \item We curate a large-scale, 3{,}400-hour multi-view dataset with over 100K trajectories and 350 daily manipulation tasks across real robots, simulation, and egocentric human videos, offering broad coverage, fine-grained annotations, and balanced supervision for training GRM.

    \item Extensive experiments validate our framework as follow: GRM achieves state-of-the-art reward assessment (over 92.8\% progress accuracy and a 0.953 Value-Order Consistency score), while on 10 simulated and 8 real-world tasks, Dopamine-RL, after one-shot GRM adaptation, raises policy success from near-zero to 95\% within 150 online rollouts (about one hour of robot interaction), with some tasks reaching 100\% success and generalizing to unseen layouts, backgrounds, and object variations.

\end{itemize}

\section{Related Work}
\label{sec:related_work}

\textbf{Reinforcement Learning for Robotic Skills.}
Reinforcement Learning (RL) has demonstrated the potential to create policies that surpass the capabilities of imitation learning~\cite{chen2025conrft, luo2025precise, lu2025vla, zhang2025reinbot, xu2024rldg, liu2505flow, li2025simplevla, zang2025rlinf, lei2025rl}, enabling the discovery of novel and robust strategies for complex, contact-rich and dexterous tasks.
Research in this area has progressed along two principal directions.
The first direction investigates various policy optimization strategies, including offline RL~\cite{luo2024multistage, luo2025precise, hu2023reboot}, online RL~\cite{liu2025can, zang2025rlinf, liu2505flow, chen2025tgrpo, li2025simplevla}, and mixed variants~\cite{chen2025conrft, mark2024policy, lei2025rl}.
The second direction explores the efficient application of RL to different model architectures, such as small fully-connected models~\cite{mark2024policy, luo2025precise}, auto-regressive models~\cite{xu2024rldg, chen2025conrft, lu2025vla}, and diffusion/flow-based models~\cite{liu2505flow, zang2025rlinf, zhang2025reinflow}.
Independent of the chosen optimization algorithm or policy architecture, a more fundamental bottleneck is to design a reward function that is effective and scalable in real-world RL, which has driven a broad shift away from manual reward engineering~\cite{peng2022deep,wu2021learning,tan2025reason,finn2016guided,van2024revisiting} toward learning-based reward models.

\textbf{Learned Process Reward Models.}
In real-world RL, a common practice is to train a success classifier as Outcome Reward Models (ORMs) to provide a binary reward signal~\cite{chen2025conrft,luo2025precise}, which renders exploration prohibitively difficult in complex, long-horizon tasks. To mitigate sparsity, recent work leverages vision–language models (VLMs) as Process Reward Models (PRMs)~\cite{ma2023liv,alakuijala2024video,ma2024gvl,zhai2025vlac,chen2025sarm}, providing denser feedback by, for example, predicting progress deltas between paired observations~\cite{zhai2025vlac} or assigning per-frame progress scores with respect to a language goal~\cite{ma2024gvl}. While several methods introduce additional structure by decomposing tasks into steps~\cite{chen2025sarm,ghasemipour2025self}, some open challenges remain (Section~\ref{sec:intro}). First, task-specific designs may limit generalization across diverse activities~\cite{ghasemipour2025self,chen2025sarm}. Second, many approaches adopt nearly uniform reward allocations, which may underweight the salience of critical sub-steps~\cite{ma2024gvl,zhai2025vlac}. In addition, current PRMs typically rely on single-view observations~\cite{ma2023liv,alakuijala2024video,ma2024gvl,zhai2025vlac,chen2025sarm}, which can impede multi-perspective state estimation and increase sensitivity to occlusions. In contrast, our method, \emph{Dopamine-Reward}, aims to address these issues by learning a general-purpose, step-aware reward model that explicitly fuses multi-view inputs, enabling a more robust and fine-grained reward estimation.

\section{Method}
\label{sec:method}

\begin{figure*}[t]
\centering
\includegraphics[width=1.0\linewidth]{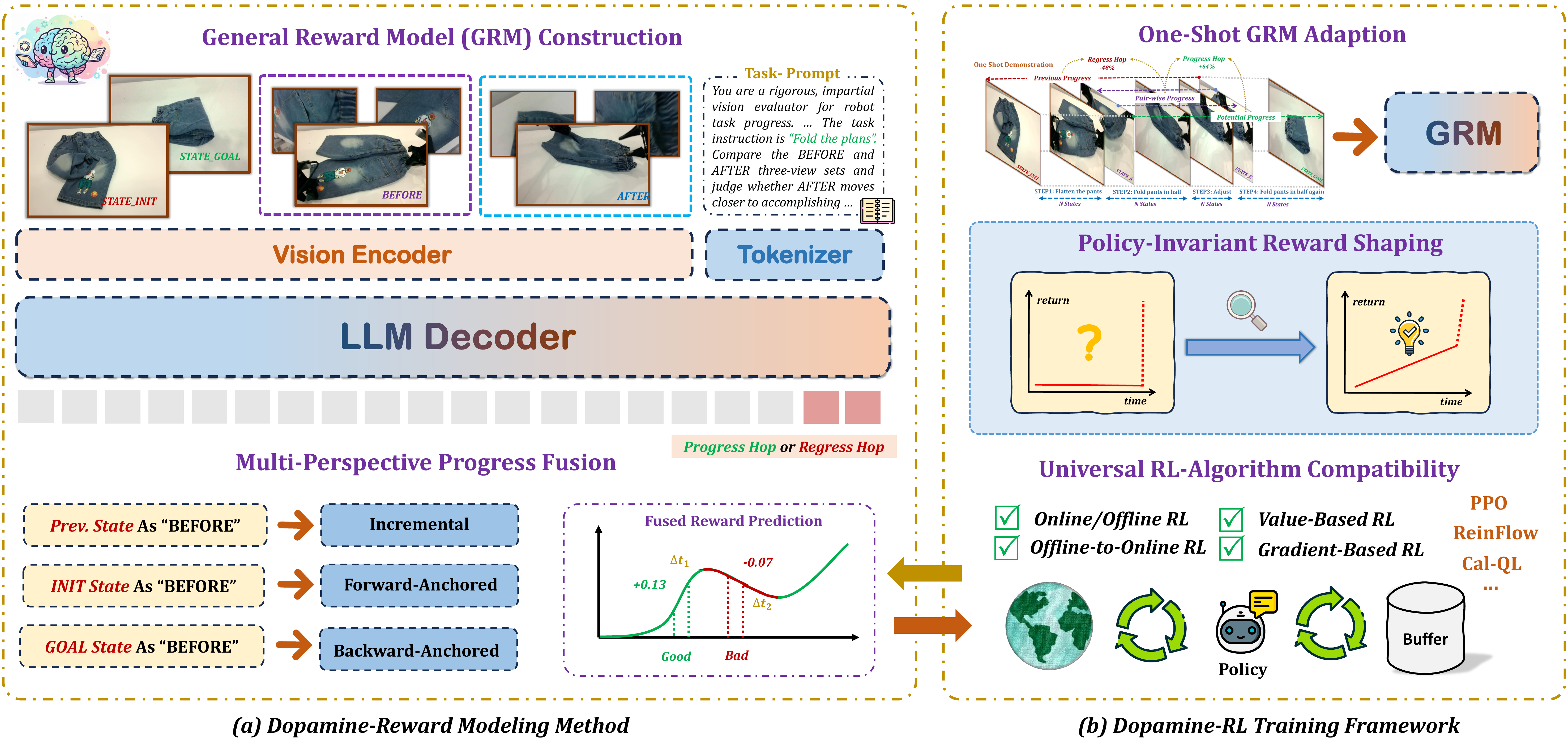}
\caption{\textbf{The overview of our method.} Our framework is composed of two core components: (a) Dopamine-Reward Modeling Method and (b) Dopamine-RL Training Framework.
\textbf{(a)} At the heart of our reward modeling is to build the General Reward Model (GRM), a vision-language model that is prompted with a task description and conditioned on multi-view images of initial, goal, ``before," and ``after" states to predict a relative progress or regress hop. To ensure a stable and accurate signal, we employ Multi-Perspective Progress Fusion, which combines incremental, forward-anchored, and backward-anchored predictions into a final fused reward.
\textbf{(b)} The Dopamine-RL framework first adapts the pre-trained GRM to a novel task using a single demonstration (One-Shot GRM Adaptation). Subsequently, it uses a theoretically-sound Policy-Invariant Reward Shaping method to convert the GRM's dense output into a reward signal that accelerates learning without altering the optimal policy. This approach is universally compatible with a wide range of RL algorithms.}
\label{fig:method}
\vspace{-0.75em}
\end{figure*}

Our approach is designed to address the core challenges in real-world robotic learning by introducing two synergistic components. First, we develop Dopamine-Reward that learns a general-purpose, step-aware process reward from multi-view inputs (\Cref{subsec:robodopamine}). Second, we propose Dopamine-RL, a robust policy learning framework built upon Dopamine-Reward, resolving the theoretical flaws in conventional reward shaping (\Cref{sec:dp-rl}).

\subsection{Dopamine-Reward Modeling Method}
\label{subsec:robodopamine}

\subsubsection{General Reward Model (GRM) Construction}
The core of our modeling method is to build the GRM, a vision-language model designed to estimate precise task progress. To ensure the model generalizes across diverse embodiments and tasks, we construct a large-scale dataset structured around relative temporal transitions. This section details the three-stage GRM training data construction pipeline, from raw video segmentation to a scientifically rigorous hop-based labeling strategy as follows:

\textit{\textbf{Step-wise task progress discretization.}}
We treat task progress itself as the supervision signal. Given raw multi-view video trajectories, we first segment each expert trajectory into sub-tasks using human-annotated multi-view keyframes $\{K_0, K_1, \dots, K_N\}$, where $K_0$ is the initial observation, $K_N$ is the final success observation, and each $K_j$ is a set of synchronized multi-view keyframes. To obtain dense supervision, we perform adaptive sampling within each segment. For a trajectory with $L$ frames per view, we set a chunk size $C$ to determine the total number of sampled points and distribute them uniformly across the $N$ segments. The number of intermediate points $m$ within segment $[K_j, K_{j+1}]$ is:
\begin{equation}
\label{eq:sampling_frames}
m = \left\lfloor \frac{1}{N} \left\lfloor \frac{L}{C} \right\rfloor \right\rfloor.
\end{equation}
This yields a sequence of states $\mathcal{S} = \{s_0, s_1, \dots, s_M\}$, 
where each state $s_i$ is a set of synchronous multi-view visual observations.
We then define the ground-truth global progress as $\Phi(s_i) = i/M$.

\textit{\textbf{Hop-based relative progress normalization.}}
A naive choice is to regress the progress gain $\Phi_{\delta}(s_p, s_q) = \Phi(s_q) - \Phi(s_p)$ between two states, but iterating such predictions accumulates error and can push the reconstructed $\Phi^{\star}(s)$ outside $[0,1]$. Instead, we introduce a hop-based formulation that learns \textit{relative-relative progress}. Each training sample is a tuple $\mathcal{D}$ containing a task description $d_{\text{task}}$, the initial state $s_0$, the goal state $s_M$, a \textit{``BEFORE''} state $s_p$, an \textit{``AFTER''} state $s_q$, and a hop label $\mathcal{H}(s_p, s_q)$ that normalizes the progress from $s_p$ to $s_q$ relative to the full task span from $s_0$ to $s_M$. Given $\Phi(s_p)$ and $\Phi(s_q)$, we define:
\begin{equation}
\label{eq:hop_calculation}
\mathcal{H}(s_p, s_q) = 
\begin{cases} 
  \dfrac{\Phi(s_q) - \Phi(s_p)}{\Phi(s_M) - \Phi(s_p)} & \text{if } q \geq p \textsc{ (progress)} \\
  \\
  \dfrac{\Phi(s_q) - \Phi(s_p)}{\Phi(s_p) - \Phi(s_0)} & \text{if } q < p \textsc{ (regress)}.
\end{cases}
\end{equation}
This dynamically scales the supervision into $[-1, 1]$: for forward progress, the change is normalized by the remaining distance to the goal; for regression, by the distance already covered from the initial state. A key theoretical advantage is that, when global progress is reconstructed by iteratively applying predicted hops, the resulting $\Phi^{\star}(s)$ is guaranteed to remain strictly within $[0,1]$. A detailed proof is provided in Appendix~\ref{app:proof1}.

\textit{\textbf{Sampling strategy and data balancing.}}
For each trajectory, we construct a balanced set of hop-based training samples. Continuous hop values are first discretized into $N_{\text{hop}}$ hop bins. The temporal distance between the \textit{``BEFORE''} state $s_p$ and \textit{``AFTER''} state $s_q$ in each pair is then chosen from $N_{\text{dis}}$ distance bins within each hop bin, yielding in total $N_{\text{hop}} \times N_{\text{dis}}$ non-trivial transitions. To reduce bias toward static segments, we further introduce an additional fraction $\alpha$ of samples explicitly labeled as zero-hop (i.e., $\mathcal{H}(s_p, s_q) = 0$), constructed by selecting pairs $(s_p, s_q)$ whose progress change is below a small threshold $\epsilon$:
\begin{equation}
|\Phi(s_q) - \Phi(s_p)| \le \epsilon.
\end{equation}

Applying this three-stage pipeline yields a dataset of 35M samples from about 3{,}400 hours of video and over 100K trajectories (see Appendix~\ref{app:data}). We train the GRM on this corpus to estimate hop-based relative progress between arbitrary state pairs, conditioned on the initial state, goal state, and task description.

\subsubsection{Multi-Perspective Progress Fusion from GRM}
\label{sec:mppf}
To mitigate error accumulation and ensure consistent accuracy, we fuse predictions based on GRM from three complementary perspectives: incremental prediction, forward-anchored prediction, and backward-anchored prediction.

\textit{Incremental Prediction} first offers a fine-grained, step-by-step assessment. Refer to \Cref{eq:hop_calculation}, the predicted global progress $\Phi_{I}^{\star}(s_t)$ is recursively computed from the preceding state's progress $\Phi^{\star}(s_{t-1})$ and the predicted hop $\mathcal{H}^\star(s_{t-1}, s_t)$. Let $\Delta\Phi^{\star}_{t-1, t}$ be the estimated progress hop:
\begin{equation}
\label{eq:incremental_pred}
\Delta\Phi^{\star}_{t-1, t} = \begin{cases} 
  [1-\Phi^{\star}(s_{t-1})] \cdot \mathcal{H}^\star & \text{if } \mathcal{H}^\star \geq 0 \\
  \Phi^{\star}(s_{t-1}) \cdot \mathcal{H}^\star & \text{if } \mathcal{H}^\star < 0.
\end{cases}
\end{equation}
The incremental progress is then calculated as follow:
\begin{equation}
\label{eq:incremental_progress}
\Phi_{I}^{\star}(s_t) = \Phi^{\star}(s_{t-1}) + \Delta\Phi^{\star}_{t-1, t},
\end{equation}
where $\Phi_{I}^{\star}(s_t)$ is be accumulated along the trajectory, initialized with $\Phi^{\star}(s_{0})=0$. While this method excels at capturing local dynamics, it is susceptible to the accumulation of prediction errors over long trajectories.
To counteract this drift, we introduce extra two global perspectives. \textit{Forward-Anchored Prediction} provides a stable global reference by anchoring to the initial state $s_\text{init}$, where progress is zero:
\begin{equation}
\label{eq:forward_pred}
\Phi_{F}^{\star}(s_t) = \mathcal{H^\star}(s_\text{init}, s_t).
\end{equation}
Conversely, \textit{Backward-Anchored Prediction} is anchored to the goal state $s_\text{goal}$, where progress is one. This approach offers high sensitivity near task completion:
\begin{equation}
\label{eq:backward_pred}
\Phi_{B}^{\star}(s_t) = 1 + \mathcal{H^\star}(s_\text{goal}, s_t).  
\end{equation}

These three methods offer complementary strengths: local precision (incremental), initial stability (forward), and goal sensitivity (backward). We fuse them via averaging to obtain a robust final progress estimate:
\begin{equation}
\label{eq:fused_pred}
\Phi^{\star}(s_t) = \frac{1}{3}\left(\Phi_{I}^{\star}(s_t) + \Phi_{F}^{\star}(s_t) + \Phi_{B}^{\star}(s_t)\right).
\end{equation}
This fusion yields a more accurate and drift-resistant signal, which is critical for the subsequent reward shaping.

\begin{figure*}[t]
    \centering
    \vspace{-0.2cm}
    \includegraphics[width=0.98\linewidth]{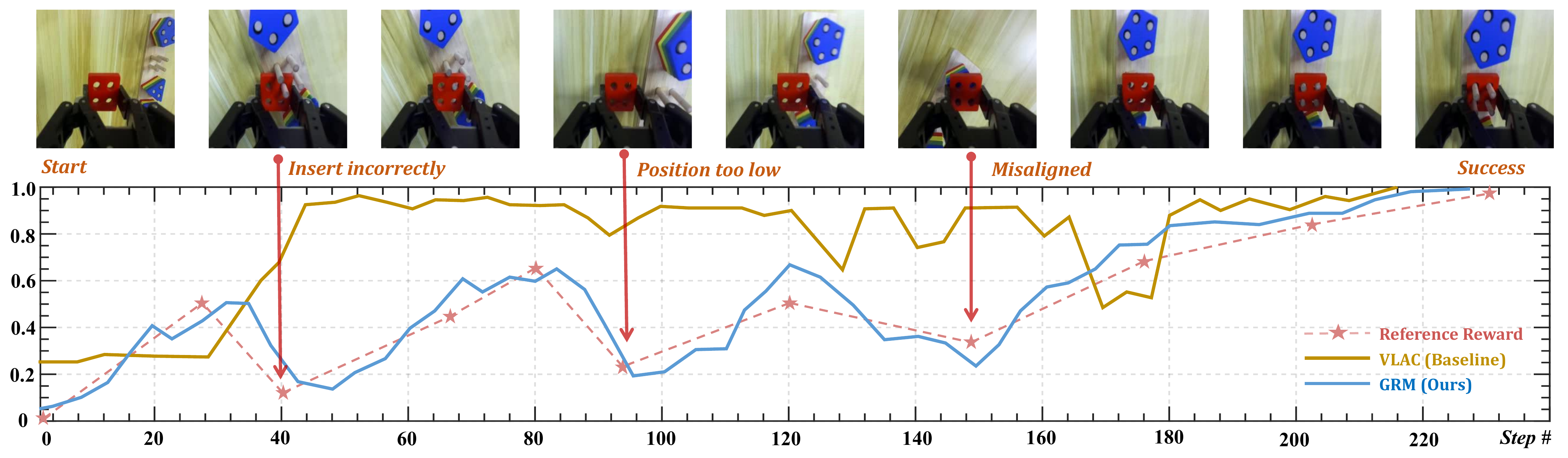}
    \caption{\textbf{Reward profiles on a challenging real-world rollout.}
    We plot the reference reward from human annotations, the VLAC baseline, and our GRM along the same trajectory. Our GRM tracks the reference signal more faithfully, sharply penalizing incorrect insertions, low positions, and misalignments, and only assigning high reward near successful task completion.}
    \label{fig:reward_profile}
\end{figure*}

\begin{figure*}[t]
    \centering
    \vspace{-0.2cm}
    \includegraphics[width=0.98\linewidth]{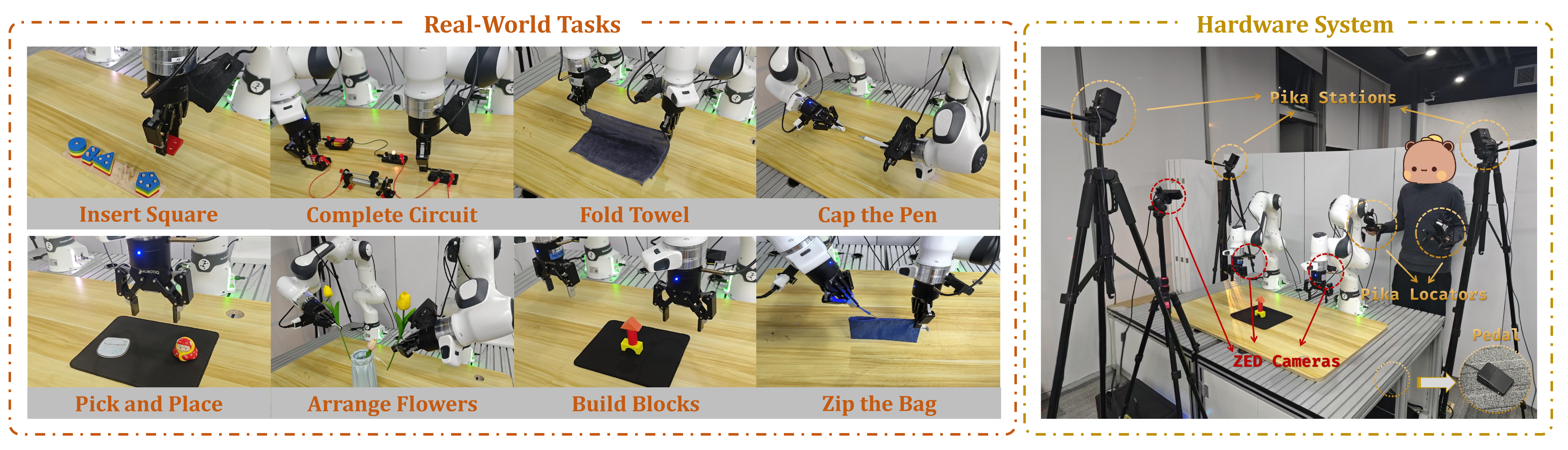}
    \caption{\textbf{Real-world tasks and hardware setup.}
    Left: eight representative long-horizon manipulation tasks used to evaluate Dopamine-Reward and Dopamine-RL, including insertion, circuit completion, folding, pick-and-place, and assembly tasks. Right: our multi-view hardware platform with the Pika teleoperation system and calibrated ZED cameras, providing synchronized wrist and third-person observations for GRM training and policy learning.}
    \label{fig:dp_hardware}
\end{figure*}

\subsubsection{Progress Consistency Checking \textit{(Optional)}}
\label{sec:robust-estimation}

While the multi-perspective fusion via averaging (\Cref{eq:fused_pred}) serves as a baseline, its naive application in online RL faces the risk of Out-of-Distribution (OOD) hallucination. Due to the inherent limitations of data coverage, it is impossible for the training set to encompass every corner of the state space. During RL, the policy inevitably explores unseen regions where the reward model may yield spurious high signals, leading to ``reward hacking.''
To address these, we propose a bi-directional consistency checking strategy that leverages consistency as a proxy for reliability, which is motivated by the observation that forward $\Phi^*_F$ and backward $\Phi^*_B$ predictions tend to exhibit significant divergence in OOD scenarios or observations, whereas they remain consistent in familiar states.

\textit{\textbf{Consistency-Aware Weighting.}}
We first define the mean estimated progress $\bar{\Phi}^*(s_t) = (\Phi^*_F(s_t) + \Phi^*_B(s_t)) / 2$. To quantify uncertainty, we calculate a normalized discrepancy metric:
\begin{equation}
    \Delta_{\text{norm}}(s_t) = \frac{|\Phi^*_B(s_t) - \Phi^*_F(s_t)|}{\bar{\Phi}^*(s_t) + \epsilon},
\end{equation}
where $\epsilon$ is a small constant for numerical stability. Normalization by $\bar{\Phi}^*$ ensures that discrepancies are penalized more heavily during the early stages (where $\Phi$ is small), as precise guidance is critical initially. We then derive a confidence weight $w_t \in (0, 1]$ using a Gaussian kernel with sensitivity $\alpha$:
\begin{equation}
    w_t = \exp\left( -\alpha \cdot (\Delta_{\text{norm}}(s_t))^2 \right).
\end{equation}

\textit{\textbf{Conservative State Update.}}
To prevent the policy from exploiting erroneous estimates in OOD scenarios, we employ a conservative update rule for the maintained progress state $\Phi^*(s_t)$ instead of \Cref{eq:fused_pred}:
\begin{equation}
    \Phi^*(s_t) = \Phi^*(s_{t-1}) + \frac{w_t}{2} \cdot \left( \bar{\Phi}^*(s_t) - \Phi^*(s_{t-1}) + \Delta\Phi^{\star}_{t-1, t} \right).
\end{equation}
This mechanism acts as a semantic filter: it ignores uncertain updates when $w_t \to 0$ (retaining $\Phi^*(s_{t-1})$) and fully trusts the estimate when consistency is high ($w_t \to 1$).

\subsection{Dopamine-RL Framework}
\label{sec:dp-rl}
Building upon Dopamine-Reward with GRM, we further introduce the Dopamine-RL framework, a reinforcement learning pipeline producing high-performance policy stimulated by Dopamine-Reward, featuring three key critical attributes: minimal downstream task effort for rapid progress alignment (\Cref{subsubsec:one-shot}), fast convergence with policy-invariant guarantees (\Cref{subsubsec:rshape}) and seamless integration with diverse RL paradigms (\Cref{subsubsec:universal}).

\subsubsection{One-shot GRM Adaptation}
\label{subsubsec:one-shot}

Dopamine-RL requires only one single human demonstration $\mathcal{D}_{\text{human}}$ to adapt the pre-trained GRM to novel or high-precision tasks, since the pre-trained GRM has already possessed a broad prior for assessing progress. Given a new task, we minimize the Mean Squared Error (MSE) between its predicted hop value, $\mathcal{H}^\star_{\omega}$, and the ground-truth, $\mathcal{H}_{\text{gt}}$:

\begin{equation}
\mathcal{L}_{\text{GRM}}(\omega) = \mathbb{E}_{(s_p, s_q) \sim \mathcal{D}_{\text{human}}}  \| \mathcal{H}^\star_{\omega} - \mathcal{H}_{\text{gt}} \|_2^2,
\label{eq:grm_sft}
\end{equation}
where $\omega$ represents the GRM's parameters, initialized by pre-trained $\text{GRM}_{\omega_0}$. After SFT, we obtain a task-adapted $\text{GRM}_{\omega_\star}$, poised for efficient reinforcement learning.
\subsubsection{Policy-Invariant Reward Shaping}
\label{subsubsec:rshape}

A straightforward approach to defining the dense process reward function for policy learning is to use the direct increment of this progress: $r(s_t, a_t, s_{t+1}) = \Phi^{\star}(s_{t+1}) - \Phi^{\star}(s_t)$. However, optimizing the standard discounted return, $J(\pi) = \mathbb{E}_{\pi}[\sum_{t=0}^{\infty}\gamma^{t}r(s_t, a_t, s_{t+1})]$, with this reward is mathematically equivalent to maximizing a different objective: $J'(\pi) \propto \mathbb{E}_{\pi}[\sum_{t=1}^{\infty}\gamma^{t-1}\Phi^{\star}(s_t) \mid s_0]$, as detailed in Appendix~\ref{app:proof2}. This transformed objective creates a perverse incentive: it encourages the agent not to complete the task, but rather to seek and maintain states with high progress values. Consequently, the resulting policy is rewarded for stagnation, preferring a safe, suboptimal state over potentially risky trajectories that lead to true task completion.
To resolve the misalignment, we formulate our GRM reward $r_{\text{GRM}}$ that adheres to three desiderata:
\begin{itemize}
    \item \textbf{Optimal policy invariance.} The optimal policy learned with $r_{\mathrm{GRM}}$ must coincide with that under the sparse gold reward $r_{\text{gold}}$ (1 at task completion, 0 otherwise), so shaping guides exploration without changing task objective.
    \item \textbf{Discount consistency:} $r_{\mathrm{GRM}}$ must be compatible with the standard exponentially discounted return and TD or Bellman updates with factor $\gamma$ under a memoryless (Markov) reward assumption (see Appendix~\ref{app:proof3}).
    \item \textbf{Locality.} At any step $t$, $r_{\mathrm{GRM}}$ is efficiently computable from the single transition $(s_t, a_t, s_{t+1})$.
\end{itemize}
Adherence to these desiderata uniquely determines the reward structure, we derive the reward from the continuous-time \textit{``discounted potential''} $e^{-\lambda t}\Phi^{\star}(s_t)$. As detailed in Appendix~\ref{app:proof4}, the natural discrete-time, single-step increment that is consistent with this continuous form is:
\begin{equation}
    F(s_t, s_{t+1}) = \gamma\Phi^{\star}(s_{t+1}) - \Phi^{\star}(s_t),
    \label{eq:shaping_term}
\end{equation}
where $\gamma = e^{-\lambda h}$. 
To enable autonomous learning on real robots without the need for continuous human monitoring, we automate the determination of the sparse outcome reward $r_{\text{gold}}$. Specifically, we consider the task completed when the estimated progress falls within a close margin of the target (i.e., $\Phi^{\star}(s_{t+1}) \ge 1 - \delta$, with $\delta=0.05$). Thus, $r_{\text{gold}}=1$ if the completion threshold is met, and $0$ otherwise. We add the shaping term $F$ to this automated gold-standard reward to define our final reward function:
\begin{equation}
    r_{\text{GRM}}(s_t, a_t, s_{t+1}) = r_{\text{gold}} + \gamma\Phi^{\star}(s_{t+1}) - \Phi^{\star}(s_t).
    \label{eq:final_reward}
\end{equation}
This form guarantees policy invariance: the cumulative discounted shaping term $F$ forms a telescoping sum that collapses to a constant boundary term depending only on the initial state $s_0$. Appendix~\ref{app:proof5} shows that the discrete-time sum and the continuous-time integral of the discounted potential's derivative converge to the same constant.
:
\begin{equation}
\begin{gathered}
    \underbrace{\sum_{t=0}^{\infty} \gamma^t (\gamma\Phi^{\star}(s_{t+1}) - \Phi^{\star}(s_t))}_{\text{Discrete PBRS Sum}}=\underbrace{-\Phi^{\star}(s_0)}_{\text{Boundary Term}} \\
    \Big\downarrow \scriptstyle \Delta t \to 0 \\
    \underbrace{\int_{0}^{\infty} \frac{d}{dt}\!\Big(e^{-\lambda t}\Phi^{\star}(s_t)\Big) dt}_{\text{Continuous Integral}} = \underbrace{-\Phi^{\star}(s_0)}_{\text{Boundary Term}}.
\end{gathered}
\label{eq:telescoping_sum}
\end{equation}
Since the shaping term telescopes to a state-dependent constant that is independent of the subsequent policy $\pi$, the shaped Q-function is simply a state-wise shift of the original one:
\begin{equation}
    Q_{\text{GRM}}^{\pi}(s,a) = Q_{\text{gold}}^{\pi}(s,a) - \Phi^{\star}(s).
    \label{eq:q_shift}
\end{equation}
The shift $-\Phi^{\star}(s)$ is identical for all actions $a$ in a given state $s$, so the optimal action remains unchanged:
\begin{equation}
    \argmax_{a} Q_{\text{GRM}}^{*}(s,a) = \argmax_{a} Q_{\text{gold}}^{*}(s,a).
    \label{eq:policy_invariance}
\end{equation}
This matches the standard Potential-Based Reward Shaping (PBRS) framework \cite{ng1999policy}, with the GRM progress $\Phi^{\star}$ serving as the potential function.

\subsubsection{Universal RL-Algorithm Compatibility}
\label{subsubsec:universal}

Dopamine-RL exhibits strong universality, seamlessly integrating with any RL algorithm, encompassing online RL, offline RL, and offline-to-online RL paradigms. It adapts effectively to both value-based methods  and gradient-based approaches.
By reshaping targeted reward functions to guide agent learning, Dopamine-RL is inherently agnostic to the specific RL algorithm employed.
Experimental results confirm this flexibility. In simulations, we deploy  under two settings: PPO\cite{ppo} (Proximal Policy Optimization) algorithm and OpenVLA-OFT\cite{openvla-oft} model, and ReinFlow\cite{reinflow} algorithm with $\pi_0$\cite{pi0} model.  exhibits excellent performance under both settings.  In real-world settings, we combine  with Cal-QL\cite{calql} (a offline-to-online Q-learning based RL algorithm) and it also delivers exceptional outcomes. Further details are shown in Appendix \ref{app:exp}.

\section{Experiments}
\label{sec:exp}

We evaluate Dopamine-Reward with GRM and Dopamine-RL on both simulation and real-world robotic platforms, covering a broad range of manipulation skills and deployment scenarios. This section summarizes our empirical findings and is organized around four questions:
\begin{itemize}
    \item \textbf{RQ1:} How accurate is the GRM at perceiving task progress compared to VLMs and existing reward models?
    \item \textbf{RQ2:} How does Dopamine-RL perform in success rate, sample efficiency, and generalization against strong BC and RL baselines?
    \item \textbf{RQ3:} How critical is Multi-Perspective Progress Fusion for final performance?
    \item \textbf{RQ4:} How important is the Dopamine-RL framework for turning reward modeling into practical policy learning?
\end{itemize}

\subsection{Accurate Task Progress Perception (RQ1)}
In this part, We assess GRM's ability to estimate task progress using two complementary protocols: video frame rank-correlation and task completion judgment.

\subsubsection{Video Frame Rank-Correlation}
\label{subsubsec:voc}

\begin{table*}[t]
\centering
\caption{\textbf{Video Frame Rank-Correlation (VOC) on Diverse Datasets.} We evaluate reward models under three temporal sampling strategies: Sparse (S), Medium (M), and Dense (D). Our GRM variants (Ours-3B and Ours-8B) consistently outperform prior work. Notably, the Ours-8B (Multi-View) model sets a new state-of-the-art across all benchmarks and sampling densities, showcasing exceptional robustness and progress understanding.}
\label{tab:voc_results}
\resizebox{\textwidth}{!}{%
\begin{tabular}{c l|ccc|ccc|ccc|ccc|ccc|ccc}
\toprule
& \multirow{2}{*}{\textbf{Dataset}} & \multicolumn{3}{c|}{\textit{GVL~\cite{ma2024gvl}}} & \multicolumn{3}{c|}{\textit{VLAC-2B~\cite{zhai2025vlac}}} & \multicolumn{3}{c|}{\textit{Ours-3B (Single-View)}} & \multicolumn{3}{c|}{\textit{Ours-3B (Multi-View)}} & \multicolumn{3}{c|}{\textit{Ours-8B (Single-View)}} & \multicolumn{3}{c}{\textit{\textbf{Ours-8B (Multi-View)}}} \\
\cmidrule(lr){3-5} \cmidrule(lr){6-8} \cmidrule(lr){9-11} \cmidrule(lr){12-14} \cmidrule(lr){15-17} \cmidrule(lr){18-20}
& & \textbf{S} & \textbf{M} & \textbf{D} & \textbf{S} & \textbf{M} & \textbf{D} & \textbf{S} & \textbf{M} & \textbf{D} & \textbf{S} & \textbf{M} & \textbf{D} & \textbf{S} & \textbf{M} & \textbf{D} & \textbf{S} & \textbf{M} & \textbf{D} \\
\midrule
\multirow{3}{*}{\rotatebox[origin=c]{90}{\small \textit{Real.}}} & DROID~\cite{khazatsky2024droid} & 0.01 & -0.30 & 0.07 & 0.66 & 0.69 & 0.50 & 0.96 & 0.95 & 0.94 & \textbf{0.99} & 0.98 & 0.97 & 0.97 & 0.96 & 0.95 & \textbf{0.99} & \textbf{0.99} & \textbf{0.98} \\
& AGIBOT-World~\cite{bu2025agibot} & 0.24 & 0.17 & 0.12 & 0.29 & 0.40 & 0.41 & 0.90 & 0.89 & 0.87 & \textbf{0.97} & 0.96 & 0.95 & 0.92 & 0.91 & 0.89 & \textbf{0.97} & \textbf{0.97} & \textbf{0.96} \\
& RoboBrain-X~\cite{FlagOpen_RoboBrainX0} & 0.32 & 0.38 & 0.33 & 0.18 & 0.13 & 0.17 & 0.85 & 0.83 & 0.81 & 0.91 & 0.89 & 0.88 & 0.87 & 0.87 & 0.83 & \textbf{0.92} & \textbf{0.91} & \textbf{0.89} \\
\midrule
\multirow{3}{*}{\rotatebox[origin=c]{90}{\small \textit{Sim.}}} & LIBERO~\cite{liu2023libero} & 0.43 & 0.37 & 0.38 & 0.19 & 0.28 & 0.41 & 0.90 & 0.86 & 0.85 & \textbf{0.95} & 0.91 & \textbf{0.92} & 0.90 & 0.88 & 0.86 & 0.94 & \textbf{0.93} & \textbf{0.92} \\
& RoboCasa~\cite{nasiriany2024robocasa} & 0.06 & -0.04 & 0.02 & 0.00 & 0.11 & 0.32 & 0.95 & 0.94 & 0.93 & 0.98 & 0.97 & 0.96 & 0.97 & 0.96 & 0.95 & \textbf{0.99} & \textbf{0.98} & \textbf{0.97} \\
& RoboTwin2.0~\cite{chen2025robotwin} & 0.28 & 0.19 & 0.10 & 0.26 & 0.35 & 0.32 & 0.90 & 0.89 & 0.87 & 0.95 & 0.94 & 0.93 & 0.92 & 0.91 & 0.89 & \textbf{0.96} & \textbf{0.95} & \textbf{0.94} \\
\midrule
\multirow{1}{*}{\small \textit{Hum.}} & EgoDex~\cite{hoque2025egodex} & 0.05 & 0.04 & -0.10 & 0.09 & 0.10 & 0.18 & \textbf{0.88} & 0.85 & 0.81 & \textbf{0.88} & 0.85 & 0.81 & \textbf{0.88} & \textbf{0.86} & \textbf{0.83} & \textbf{0.88} & \textbf{0.86} & \textbf{0.83} \\
\midrule
& \textbf{Average} 
& 0.20 & 0.12 & 0.13   
& 0.24 & 0.29 & 0.33   
& 0.91 & 0.89 & 0.87   
& 0.96 & 0.94 & 0.93   
& 0.92 & 0.91 & 0.89   
& \textbf{0.96} & \textbf{0.96} & \textbf{0.94} \\ 
\bottomrule
\end{tabular}%
}
\end{table*}

To quantitatively assess task progress perception, we follow the evaluation methodology of GVL~\cite{ma2024gvl} and measure the Value-Order Correlation (VOC) between the GRM's predicted progress and the ground-truth chronological order of shuffled frames. A higher VOC score ([-1, 1]) indicates a better understanding of temporal progress. We evaluate on a diverse suite of eight datasets spanning real-world robotics (DROID~\cite{khazatsky2024droid}, AGIBOT-World~\cite{bu2025agibot}, RoboBrain-X~\cite{FlagOpen_RoboBrainX0}), simulation (Libero~\cite{liu2023libero}, RoboCasa~\cite{nasiriany2024robocasa}, RoboTwin2.0~\cite{chen2025robotwin}), and egocentric human manipulation (EgoDex~\cite{hoque2025egodex}). To test robustness to temporal granularity, we test under three distinct sampling strategies: \textit{Sparse (S)} using only major keyframes, \textit{Medium (M)} using uniform samples between keyframes, and \textit{Dense (D)} using uniform samples across the entire trajectory. We compare our multi-view and single-view GRM against four state-of-the-art reward models: GVL~\cite{ma2024gvl}, and VLAC~\cite{zhai2025vlac}.
As shown in Table~\ref{tab:voc_results}, our multi-view GRM consistently achieves the highest VOC scores across all seven datasets and all sampling strategies. The performance of baseline models tends to degrade as sampling becomes denser, indicating a struggle with fine-grained temporal distinctions. In contrast, our model maintains exceptionally high performance, highlighting the robustness of our hop-based learning formulation and multi-perspective fusion. The performance gap is most significant in complex, long-horizon tasks (e.g., LIBERO~\cite{liu2023libero}, RoboBrain-X~\cite{FlagOpen_RoboBrainX0}), where our model's ability to accurately contextualize progress is paramount.

\begin{table}[h!]
\centering
\caption{\textbf{Task Completion Classification Accuracy (as successes out of 60).} Our GRM more accurately classifies the final outcomes of robot rollouts compared to both specialized reward models and large generalist models.}
\label{tab:completion_judgment}
\scalebox{0.78}{
\begin{tabular}{c l|ccc|c}
\toprule
& \textbf{Method} & \textbf{Stacking} & \textbf{Folding} & \textbf{Clearing} & \textbf{Average} \\
\midrule
\multirow{4}{*}{\rotatebox[origin=c]{90}{\small \textit{VLMs}}} 
& Gemini-2.5-Pro~\cite{comanici2025gemini} & $50/60$ & $45/60$ & $51/60$ & $81.1\%$ \\
& GPT-5~\cite{openai_gpt5_system_card_pdf} & $51/60$ & $48/60$ & $52/60$ & $83.9\%$ \\
& Qwen3-VL~\cite{QwenLM_Qwen3_VL} & $43/60$ & $41/60$ & $43/60$ & $76.7\%$ \\
& RoboBrain 2.0~\cite{team2025robobrain} & $38/60$ & $35/60$ & $41/60$ & $61.7\%$ \\
\midrule
\multirow{2}{*}{\rotatebox[origin=c]{90}{\small \textit{RMs}}} 
& GVL~\cite{ma2024gvl} & $25/60$ & $27/60$ & $15/60$ & $37.2\%$ \\
& VLAC-2B~\cite{zhai2025vlac} & $19/60$ & $21/60$ & $21/60$ & $33.9\%$ \\
\midrule
\multicolumn{2}{c|}{\textit{Ours-8B (Single-View)}} 
& $50/60$ & $50/60$ & $51/60$ & $83.9\%$ \\
\multicolumn{2}{c|}{\textit{\textbf{Ours-8B (Multi-View)}}} 
& $\mathbf{56/60}$ & $\mathbf{54/60}$ & $\mathbf{57/60}$ & $\mathbf{92.8\%}$ \\
\bottomrule
\end{tabular}}
\end{table}

\subsubsection{Task Completion Judgment}

To assess the GRM's ability to make high-level judgments about task outcomes, we follow the protocol from SARM~\cite{chen2025sarm}. We collect 60 real-world rollouts for each of three tasks (stacking blocks, folding T-shirt, clearing desktop), with 20 successful (SE), 20 partially successful (PSE), and 20 failed (FE) episodes. We evaluate classification accuracy against reward model baselines (GVL~\cite{ma2024gvl}, VLAC~\cite{zhai2025vlac}) and generalist vision-language models (Gemini-2.5-Pro~\cite{comanici2025gemini}, GPT-5~\cite{openai_gpt5_system_card_pdf}, Qwen3-VL-8B~\cite{QwenLM_Qwen3_VL}, RoboBrain 2.0-8B~\cite{team2025robobrain,ji2025robobrain}). More evaluation settings are shown in Appendix~\ref{app:grm-exp}. Results in Table~\ref{tab:completion_judgment} shows: \textit{\textbf{(1) Superiority over Generalist VLMs:}} While large models like GPT-5~\cite{openai_gpt5_system_card_pdf} and Gemini-2.5-Pro~\cite{comanici2025gemini} achieve respectable accuracy ($\sim$83\%), they often struggle with spatial precision—misclassifying ``near-misses" (PSE) as successes. Our GRM-8B (Multi-View) significantly outperforms them (+9\% vs GPT-5), demonstrating that domain-specific training on progress data is more effective than zero-shot reasoning.
\textit{\textbf{(2) Failure of Single-View PRMs:}} Existing reward models like GVL~\cite{ma2024gvl} and VLAC~\cite{zhai2025vlac} perform poorly (accuracy $<40\%$). This is primarily because single-view models lose track of objects during occlusion (\textit{e.g.,} hand covering the block during stacking), leading to noisy progress curves that fail the stability check in \Cref{eq:judgment_logic}.
\textit{\textbf{(3) Impact of Multi-View Fusion:}} The gap between our Single-View (83.9\%) and Multi-View (92.8\%) variants highlights the critical role of perceptual robustness. The multi-view fusion ensures that if one view is occluded, the model can still verify progress via alternative angles, correctly distinguishing between a completed task (SE) and a stalled one (PSE/FE).

\begin{table}[h!]
\centering
\caption{\textbf{Policy Performance and Sample Efficiency.} Dopamine-RL achieves significantly higher performance with fewer human demonstrations. Sample efficiency is measured by episodes needed to reach 80\% of the final success rate (lower is better).}
\label{tab:main_results}
\scalebox{0.82}{
\begin{tabular}{l|cc|cc}
\toprule
\multirow{2}{*}{\textbf{Method}} & \multicolumn{2}{c|}{\textbf{Simulation} (10 Tasks)} & \multicolumn{2}{c}{\textbf{Real-World} (8 Tasks)} \\
\cmidrule(lr){2-3} \cmidrule(lr){4-5}
 & \textbf{SR (\%)} & \textbf{Rollout (\#)} & \textbf{SR (\%)} & \textbf{Rollout (\#)} \\
\midrule
BC (50 demos) & 31.5 & -- & 9.8 & -- \\
RL + Sparse & 79.9 & 560 & 68.0 & 183 \\
\midrule
\textbf{Dopamine-RL} & \textbf{81.0} & \textbf{395} & \textbf{95.2} & \textbf{150} \\

\bottomrule
\end{tabular}}
\end{table}

\subsection{Performance, Efficiency, Generalization (RQ2)}
We now evaluate the Dopamine-RL framework across 10 simulation tasks (from LIBERO~\cite{liu2023libero} and RoboTwin2.0~\cite{chen2025robotwin}) and 8 real-world tasks, whose task setups and hardware platform are illustrated in Figure~\ref{fig:dp_hardware}. 
In our simulation experiments, Dopamine-RL is evaluated under two distinct configurations: one leveraging the PPO~\cite{ppo} (Proximal Policy Optimization) algorithm alongside the OpenVLA-OFT~\cite{openvla-oft} model, and the other integrating the ReinFlow~\cite{reinflow} algorithm with the $\pi_0$~\cite{pi0} model. For real-world implementations, we pair Dopamine-RL with Cal-QL~\cite{calql} and we employ a Human-in-the-Loop setup where we use just one single human demonstrations to adapt the GRM. We compare against strong baselines: Behavioral Cloning (BC) on 50 demos, and Proximal Policy Optimization~\cite{ppo} (PPO) using a sparse reward in simulation and ConRFT~\cite{conrft} for real-world settings.

As shown in Table~\ref{tab:main_results}, Dopamine-RL significantly outperforms all baselines in both final success rate and sample efficiency. The dense and accurate rewards from our GRM enables rapid and stable learning, achieving high performance with far fewer environment interactions.
To test generalization, we evaluate the final policies under In-Distribution (ID) and Out-of-Distribution (OOD) conditions, where OOD settings include changes to object properties (Object Change), workspace layout (Layout Change), and background visuals (Background Change).
Table~\ref{tab:ood_results_transposed} presents a detailed breakdown of this analysis. The results show that while both methods experience a performance drop when faced with distribution shifts, the Average Relative Drop ($\Delta$) is significantly more pronounced for the BC baseline (50-60\% degradation). In contrast, our Dopamine-RL framework maintains a much higher proportion of its original performance, with a relative drop of only 8-20\%. This quantitatively demonstrates that our policy has learned a more robust and generalizable understanding of the task semantics, successfully mitigating the overfitting to superficial visual features that severely impacts the BC baseline.

\begin{table}[t]
\centering
\caption{\textbf{Generalization Performance Breakdown: ID vs. OOD.} The table compares success counts (out of 20 trials) for Behavioral Cloning (BC) and our framework (Ours). The final row, Avg. Relative Drop ($\Delta$), quantifies the average relative success rate drop from ID performance when tested on OOD settings.}
\label{tab:ood_results_transposed}
\scalebox{0.81}{
\begin{tabular}{c l|cc|cc|cc}
\toprule
& \multirow{2}{*}{\textbf{Condition}} & \multicolumn{2}{c|}{\textbf{Insert Square}} & \multicolumn{2}{c|}{\textbf{Circuit}} & \multicolumn{2}{c}{\textbf{Cap Pen}} \\
\cmidrule(lr){3-4} \cmidrule(lr){5-6} \cmidrule(lr){7-8}
&  & \textbf{BC} & \textbf{Ours} & \textbf{BC} & \textbf{Ours} & \textbf{BC} & \textbf{Ours} \\
\midrule
\multicolumn{2}{c|}{Original (ID) $\uparrow$} & 7/20 & \textbf{19/20} & 5/20 & \textbf{20/20} & 8/20 & \textbf{19/20} \\
\midrule
\multirow{3}{*}{\rotatebox[origin=c]{90}{\small \textit{OOD}}} & Object $\uparrow$ & 4/20 & \textbf{15/20} & 3/20 & \textbf{17/20} & 5/20 & \textbf{17/20} \\
& Layout $\uparrow$ & 2/20 & \textbf{15/20} & 1/20 & \textbf{19/20} & 3/20 & \textbf{15/20} \\
& Background $\uparrow$ & 3/20 & \textbf{16/20} & 2/20 & \textbf{19/20} & 4/20 & \textbf{16/20} \\
\midrule
\multicolumn{2}{c|}{Avg. Drop ($\Delta$ \%) $\downarrow$} & 57.1 & \textbf{19.3} & 60.0 & \textbf{8.3} & 50.0 & \textbf{15.8} \\
\bottomrule
\end{tabular}}
\end{table}

\subsection{Ablation Studies (RQ3 \& RQ4)}
Finally, we conduct a series of ablation studies on a representative subset of three real-world tasks to validate the key design choices in the Dopamine-Reward framework.
The results in Table~\ref{tab:ablation_results} confirm our hypotheses.

\begin{table}[t]
\centering
\caption{\textbf{Ablation Study Results (Average Success Rate \%).} Each component of the Dopamine-Reward framework is shown to be critical for achieving maximum performance.}
\label{tab:ablation_results}
\scalebox{0.78}{
\begin{tabular}{l|c|c}
\toprule
\textbf{Method Variation} & \textbf{Success Rate} & \textbf{$\Delta$ from Full} \\
\midrule
\textbf{Full Framework (Dopamine-RL)} & \textbf{85.0} & \textbf{--} \\
\midrule
\rowcolor[HTML]{F2F2F2} \multicolumn{3}{l}{\textit{Ablations for RQ3 (Core Components)}} \\ \midrule
w/o Fusion (Incremental Only) & 70.0 & -15.0 \\
w/o Fusion (Forward-Anchored Only) & 65.7 & -19.3 \\
w/o Fusion (Backward-Anchored Only) & 62.5 & -22.5 \\
\midrule
\rowcolor[HTML]{F2F2F2} \multicolumn{3}{l}{\textit{Ablations for RQ4 (Dopamine-RL Framework)}} \\ \midrule
w/o Policy-Invariant Shaping & 41.3 & -43.7 \\
w/o One-shot adaption & 63.2 & -21.8 \\
\bottomrule
\end{tabular}}
\end{table}

For \textbf{\textit{RQ3}}, we ablate the Multi-Perspective Progress Fusion in Dopamine-Reward. Removing fusion and relying on a single progress estimator consistently hurts performance: the incremental-only, forward-anchored-only, and backward-anchored-only variants incur $15.0\%$, $19.3\%$, and $22.5\%$ absolute drops, respectively. The incremental-only variant is particularly vulnerable to error drift over long horizons, confirming the importance of combining local and global progress perspectives.

For \textbf{\textit{RQ4}}, the importance of the Dopamine-RL is evident. 
Removing policy-invariant reward shaping leads to a massive performance drop of 43.7\%. The agent learns to reach ``good-enough'' states and stagnates, failing to complete the tasks, which confirms the ``semantic trap'' discussed in~\Cref{sec:method}.
Besides, relying solely on zero-shot GRM, it occasionally provides incorrect rewards for corner cases in out-of-distribution (OOD) tasks, such as assigning positive rewards to poor actions and negative rewards to good ones. This hinders the convergence of the policy, resulting in a 21.8\% drop in success rate.

\section{Conclusion}
\label{sec:conclusion}

In this work, as named Robo-Dopamine, we tackled the critical challenges of reward design in real-world robotics by introducing Dopamine-Reward, a novel approach for learning a general-purpose, step-aware reward model from multi-view inputs. Our core contribution, the General Reward Model (GRM), is trained on over 3,400 hours of diverse data processed via Dopamine-Reward and leverages Multi-Perspective Progress Fusion to overcome perceptual limitations like occlusion. Building upon this, our Dopamine-RL framework employs a theoretically-grounded, Policy-Invariant Reward Shaping method, which provides dense guidance to accelerate learning without altering the optimal policy, thereby systematically avoiding the common ``semantic trap". Extensive experiments on diverse tasks validate our approach, demonstrating state-of-the-art reward accuracy and remarkable sample efficiency, with policies improving success rates from nearly-zero to $\sim$95\% in an average of only $\sim$150 interaction rollouts while exhibiting strong generalization. By combining a robust multi-view reward model with a principled RL framework, our work presents a scalable recipe for enabling embodied agents to achieve continuous self-improvement and master complex manipulation tasks far beyond their initial demonstrations. In the future, we plan to expand our work in four potential directions, detailed in Appendix.~\ref{future}.

\clearpage
{
    \small
    \bibliographystyle{ieeenat_fullname}
    \bibliography{main}
}
\clearpage
\maketitlesupplementary
\newpage
\appendix
\section*{Appendix}
This supplementary material provides comprehensive details regarding the proposed method, Dopamine-Reward, and the policy learning framework, Dopamine-RL, along with additional experimental results omitted from the main manuscript due to space constraints.
Section~\ref{app:prof_all} provides rigorous theoretical proofs for the bounded global progress, the existence of the ``semantic trap,'' and the optimal policy invariance, etc.
Section~\ref{app:data} elaborates on the composition, statistics, and sampling strategies of our 35M-sample training dataset.
Section~\ref{app:exp} outlines detailed experimental setups for GRM, simulation and real-world evaluations.
Section~\ref{app:vis} presents additional qualitative visualizations to demonstrate the effectiveness of our General Reward Model (GRM).
Finally, Section~\ref{future} discusses limitations and potential future research directions.

\section{Proof}
\label{app:prof_all}
\subsection{Proof of Bounded Global Progress (Proof 1)}
\label{app:proof1}

In this subsection, we provide a formal proof that iteratively applying the predicted relative progress hops guarantees that the reconstructed global progress $\Phi^{\star}(s)$ remains strictly within the bounds $[0, 1]$, provided that the initial state is bounded and the model predictions lie within $[-1, 1]$.

First, we define the general recursive update rule. Based on the definition of the hop label $\mathcal{H}(s_p, s_q)$ in Equation~\ref{eq:hop_calculation}, we derive the recursive update rule for estimating the global progress of the next state $\Phi^{\star}(s_t)$ given the current state $\Phi^{\star}(s_{t-1})$ and the predicted hop $H = \mathcal{H}(s_{t-1}, s_t)$. We assume the normalization where $\Phi(s_0)=0$ and $\Phi(s_M)=1$. Rearranging the equation, the update rule is:
\begin{equation}
\label{eq:update_rule}
\Phi^{\star}(s_t) = 
\begin{cases} 
    \Phi^{\star}(s_{t-1}) + H \cdot [1 - \Phi^{\star}(s_{t-1})] & \text{if } H \ge 0 \\
    \Phi^{\star}(s_{t-1}) + H \cdot \Phi^{\star}(s_{t-1}) & \text{if } H < 0
\end{cases}
\end{equation}

Given that the initial progress $\Phi^{\star}(s_0) = 0$ and the predicted hop $H \in [-1, 1]$, the reconstructed global progress $\Phi^{\star}(s_t)$ satisfies $\Phi^{\star}(s_t) \in [0, 1]$ for all steps $t$.

We proceed by mathematical induction as follow:
\textit{(1) Base Case ($t=0$):}
By definition, $\Phi^{\star}(s_0) = 0$, which satisfies $0 \in [0, 1]$.
\textit{(2) Inductive Step:}
Assume that for step $t-1$, the hypothesis holds: $0 \le \Phi^{\star}(s_{t-1}) \le 1$.
Let $G = \Phi^{\star}(s_{t-1})$ for brevity, where $G \in [0, 1]$. We analyze the next state $\Phi^{\star}(s_t)$ under two cases (\textit{i.e.,} Positive Hop and Negative Hop) based on the sign of the predicted hop $H$.

\begin{itemize}
    \item \textbf{\textit{Case 1: Positive Hop (Progress), $0 \le H \le 1$.}} \\
    From~\Cref{eq:update_rule}, the update is written as:
    \begin{equation}
        \Phi^{\star}(s_t) = G + H(1 - G)
    \end{equation}
    Rearranging terms to view this as a convex combination:
    \begin{equation}
        \Phi^{\star}(s_t) = H + G(1 - H)
    \end{equation}
    
    \textit{Lower Bound:} Since $G \ge 0$, $H \ge 0$, and $(1-H) \ge 0$, it follows that $\Phi^{\star}(s_t) \ge 0$.
    
    \textit{Upper Bound:} Since $G \le 1$, we substitute the maximum value of $G$:
    \begin{align*}
        \Phi^{\star}(s_t) &= H + G(1 - H) \\
        &\le H + 1 \cdot (1 - H) \\
        &= H + 1 - H \\
        &= 1
    \end{align*}
    Thus, $0 \le \Phi^{\star}(s_t) \le 1$ when $H \ge 0$.

    \item \textit{\textbf{Case 2: Negative Hop (Regress), $-1 \le H < 0$.}} \\
    From~\Cref{eq:update_rule}, the update is:
    \begin{equation}
        \Phi^{\star}(s_t) = G + H \cdot G = G(1 + H)
    \end{equation}
    
    \textit{Lower Bound:} Since $H \in [-1, 0)$, the term $(1 + H) \ge 0$. Since $G \ge 0$, the product $G(1+H) \ge 0$.
    
    \textit{Upper Bound:} Since $H < 0$, the term $(1 + H) < 1$. Combining this with $G \le 1$:
    \begin{equation*}
        \Phi^{\star}(s_t) = G(1 + H) \le 1 \cdot (1) = 1
    \end{equation*}
    Thus, $0 \le \Phi^{\star}(s_t) \le 1$ when $H < 0$.
\end{itemize}

\noindent\textbf{Conclusion.} Since the property holds for the base case and is preserved in both update scenarios during the inductive step, we conclude that $\Phi^{\star}(s_t) \in [0, 1]$ for all $t$.

\subsection{Proof of the Semantic Trap (Proof 2)}
\label{app:proof2}

In this subsection, we provide a theoretical derivation demonstrating why a naive dense reward formulation, defined as the direct increment of progress $r(s_t, a_t, s_{t+1}) = \Phi(s_{t+1}) - \Phi(s_t)$, fundamentally alters the reinforcement learning objective, leading to the ``Semantic Trap'' described in the main text.

Consider the standard objective in reinforcement learning, which is to maximize the expected discounted return with a discount factor $\gamma \in (0, 1)$:
\begin{equation}
    J(\pi) = \mathbb{E}_{\pi} \left[ \sum_{t=0}^{\infty} \gamma^t r(s_t, a_t, s_{t+1}) \bigg| s_0 \right].
\end{equation}
Substituting the naive progress-difference reward $r(s_t, a_t, s_{t+1}) = \Phi(s_{t+1}) - \Phi(s_t)$ into the cumulative return for a finite horizon $T$:
\begin{equation}
    G_T = \sum_{t=0}^{T-1} \gamma^t [\Phi(s_{t+1}) - \Phi(s_t)].
\end{equation}
We can split the summation into two distinct terms:
\begin{equation}
    G_T = \sum_{t=0}^{T-1} \gamma^t \Phi(s_{t+1}) - \sum_{t=0}^{T-1} \gamma^t \Phi(s_t).
\end{equation}
By applying a variable substitution $k = t+1$ to the first term, we rewrite it as $\sum_{k=1}^{T} \gamma^{k-1} \Phi(s_k)$. The second term can be expanded as $\Phi(s_0) + \sum_{t=1}^{T-1} \gamma^t \Phi(s_t)$. Substituting these back into the expression for $G_T$ yields:
\begin{equation}
\begin{split}
    G_T = &\left[ \sum_{t=1}^{T-1} \gamma^{t-1} \Phi(s_t) + \gamma^{T-1} \Phi(s_T) \right] - \\
    & \left[ \Phi(s_0) + \sum_{t=1}^{T-1} \gamma^t \Phi(s_t) \right].
\end{split}
\end{equation}
We now group the terms for each time step $t \in [1, T-1]$:
\begin{align}
    G_T &= -\Phi(s_0) + \sum_{t=1}^{T-1} (\gamma^{t-1} - \gamma^t) \Phi(s_t) + \gamma^{T-1} \Phi(s_T) \\
    &= -\Phi(s_0) + (1-\gamma) \sum_{t=1}^{T-1} \gamma^{t-1} \Phi(s_t) + \gamma^{T-1} \Phi(s_T).
\end{align}
Since the progress metric $\Phi(s)$ is bounded within $[0, 1]$ and $\gamma \in (0, 1)$, the term $\gamma^{T-1} \Phi(s_T)$ vanishes as $T \to \infty$. The infinite horizon return converges to:
\begin{equation}
    G = \lim_{T \to \infty} G_T = -\Phi(s_0) + (1-\gamma) \sum_{t=1}^{\infty} \gamma^{t-1} \Phi(s_t).
\end{equation}
Because $\Phi(s_0)$ is a constant with respect to the policy $\pi$, maximizing the expected return is equivalent to:
\begin{equation}
    \arg\max_{\pi} J(\pi) \propto \arg\max_{\pi} \mathbb{E}_{\pi} \left[ \sum_{t=1}^{\infty} \gamma^{t-1} \Phi(s_t) \bigg| s_0 \right].
\end{equation}
\textbf{Conclusion:} The optimization objective implicitly shifts from maximizing the \textit{change} in progress to maximizing the \textit{accumulated value} of progress states over time. This creates a perverse incentive where the agent is encouraged to reach a high-progress state quickly and stagnate there to accumulate rewards at each step, rather than completing the task. This theoretical result confirms the existence of the ``Semantic Trap.''
\subsection{Derivation of Exponential Discounting from Time-Consistency (Proof 3)}
\label{app:proof3}

In this subsection, we justify the use of the exponential discount factor $\gamma = e^{-\lambda h}$. We start from the principle of \textit{time-consistency} (or memorylessness). Let $D(t): \mathbb{R}_{\ge 0} \to (0, 1]$ be a discount function. The memoryless property implies that the relative discount factor for an additional delay $\Delta$ should not depend on how much time $\tau$ has already passed:
\begin{equation}
    \frac{D(\tau + \Delta)}{D(\tau)} = D(\Delta), \quad \forall \tau, \Delta \ge 0.
\end{equation}
Setting $\tau=t$ and $\Delta=s$, this yields the Cauchy functional equation:
\begin{equation}
    D(t+s) = D(t)D(s).
\end{equation}
Transforming to the logarithmic domain with $\phi(t) = \ln D(t)$, we obtain the linear equation $\phi(t+s) = \phi(t) + \phi(s)$, the unique continuous solution of which is $\phi(t) = -\lambda t$ for some constant $\lambda \ge 0$. Thus, the discount function must take the exponential form:
\begin{equation}
    D(t) = e^{-\lambda t}.
\end{equation}
In the discrete setting with time step $h$, this corresponds to the discount factor $\gamma = D(h) = e^{-\lambda h}$. This theoretical foundation ensures that our reward formulation is robust to variations in control frequency.

\subsection{Consistency of Reward Shaping in Continuous and Discrete Time (Proof 4)}
\label{app:proof4}

In this subsection, we derive the continuous-time counterpart of our discrete potential-based reward shaping term $\gamma \Phi(s_{t+1}) - \Phi(s_t)$. We demonstrate that the discrete shaping term is mathematically equivalent to the first-order Euler discretization of a specific differential equation. Furthermore, we show that its cumulative sum converges to a boundary term that ensures policy invariance.

\subsubsection{Notation and Definitions}
Let $h = \Delta t$ denote the discretization time step. We map the discrete time steps $t, t+1$ to continuous time moments $t, t+h$. Let $s(t)$ denote the state trajectory in continuous time.
\begin{itemize}
    \item The relationship between the continuous discount rate $\lambda > 0$ and the discrete discount factor $\gamma$ is defined as:
    \begin{equation}
        \gamma = \exp(-\lambda h).
    \end{equation}
    \item Let $\Phi(t) \coloneqq \Phi(s(t))$ denote the potential value along the trajectory. Its total time derivative is:
    \begin{equation}
        \dot{\Phi}(t) = \frac{d}{dt}\Phi(s(t)).
    \end{equation}
\end{itemize}

\subsubsection{First-Order Taylor Expansion}
We perform a first-order Taylor expansion for both the potential function $\Phi(s(t+h))$ and the discount factor $\gamma$ with respect to the step size $h$:
\begin{align}
    \Phi(s(t+h)) &= \Phi(s(t)) + h\,\dot{\Phi}(t) + O(h^2), \\
    \gamma = e^{-\lambda h} &= 1 - \lambda h + O(h^2).
\end{align}

\subsubsection{Derivation of the Instantaneous Shaping Term}
Substituting the expansions into the discrete shaping term $\gamma \Phi(s_{t+1}) - \Phi(s_t)$ (where $s_{t+1} \equiv s(t+h)$):
\begin{equation}
\begin{split}
    &\gamma \Phi(s(t+h)) - \Phi(s(t)) \\
    &= (1 - \lambda h)(\Phi(t) + h\,\dot{\Phi}(t)) - \Phi(t) + O(h^2) \\
    &= \left( \Phi(t) + h\dot{\Phi}(t) - \lambda h \Phi(t) - \lambda h^2 \dot{\Phi}(t) \right) \\
    &\quad - \Phi(t) + O(h^2) \\
    &= h \left( \dot{\Phi}(t) - \lambda \Phi(t) \right) + O(h^2).
\end{split}
\end{equation}
Dividing by $h$ and taking the limit as $h \to 0$ yields the \textit{instantaneous density} of the reward shaping:
\begin{equation}
    \lim_{h \to 0} \frac{\gamma \Phi(s(t+h)) - \Phi(s(t))}{h} = \dot{\Phi}(t) - \lambda \Phi(t).
\end{equation}
This result indicates that the single-step potential-based reward is, to the first order, the rectangular integration of $\dot{\Phi}(t) - \lambda \Phi(t)$ over the interval $[t, t+h]$.

\subsubsection{From Cumulative Sum to Integral Form}
We now consider the cumulative discounted sum of the shaping reward over a horizon $N$. Let $t_k = k \cdot h$. The discount factor at step $k$ is $\gamma^k = (e^{-\lambda h})^k = e^{-\lambda t_k}$.
The discrete cumulative sum is:
\begin{equation}
    \sum_{k=0}^{N-1} \gamma^k \left[ \gamma \Phi(s_{k+1}) - \Phi(s_k) \right].
\end{equation}
Substituting the first-order approximation derived above:
\begin{equation}
\begin{split}
    &\sum_{k=0}^{N-1} e^{-\lambda t_k} \left[ h (\dot{\Phi}(t_k) - \lambda \Phi(t_k)) + O(h^2) \right] \\
    &= \sum_{k=0}^{N-1} h \cdot e^{-\lambda t_k} (\dot{\Phi}(t_k) - \lambda \Phi(t_k)) + O(h).
\end{split}
\end{equation}
As $h \to 0$, this Riemann sum converges to the definite integral:
\begin{equation}
    \xrightarrow{h \to 0} \int_{0}^{T} e^{-\lambda t} \left( \dot{\Phi}(t) - \lambda \Phi(t) \right) dt.
\end{equation}

\subsubsection{Boundary Terms and Consistency}
Using the product rule for differentiation, we observe that the integrand is exactly the total derivative of the discounted potential:
\begin{equation}
    \frac{d}{dt} \left( e^{-\lambda t} \Phi(t) \right) = e^{-\lambda t} \dot{\Phi}(t) - \lambda e^{-\lambda t} \Phi(t) = e^{-\lambda t} (\dot{\Phi}(t) - \lambda \Phi(t)).
\end{equation}
Thus, the integral can be evaluated analytically:
\begin{equation}
\begin{split}
    \int_{0}^{T} \frac{d}{dt} \left( e^{-\lambda t} \Phi(t) \right) dt &= \left[ e^{-\lambda t} \Phi(s(t)) \right]_{0}^{T} \\
    &= e^{-\lambda T} \Phi(s(T)) - \Phi(s(0)).
\end{split}
\end{equation}
Assuming $\Phi(s)$ is bounded, as $T \to \infty$, the term $e^{-\lambda T} \Phi(s(T))$ vanishes. The cumulative shaping reward simplifies to a constant boundary term:
\begin{equation}
    \lim_{T \to \infty} \int_{0}^{T} e^{-\lambda t} (\dot{\Phi} - \lambda \Phi) dt = - \Phi(s(0)).
\end{equation}
This confirms that in continuous time, the total shaping reward depends only on the initial state, preserving policy invariance.

\subsubsection{The ODE / Euler Method Perspective}
Finally, we provide an intuitive interpretation using Ordinary Differential Equations (ODEs). Let us define the \textit{discounted potential} as a state variable $y(t)$:
\begin{equation}
    y(t) \coloneqq e^{-\lambda t} \Phi(s(t)).
\end{equation}
The dynamics of $y(t)$ are governed by:
\begin{equation}
    \frac{dy}{dt} = e^{-\lambda t} (\dot{\Phi}(t) - \lambda \Phi(t)).
\end{equation}
If we apply the \textbf{Forward Euler method} to solve this ODE numerically at discrete steps $t_k$ with step size $h$:
\begin{equation}
    y(t_{k+1}) \approx y(t_k) + h \cdot \frac{dy}{dt}\bigg|_{t_k}.
\end{equation}
Substituting the definitions back:
\begin{equation}
    \begin{split}
    e^{-\lambda(t_k+h)} \Phi(s_{k+1}) \approx &e^{-\lambda t_k} \Phi(s_k) \\
    &+ h \cdot e^{-\lambda t_k} (\dot{\Phi}(t_k) - \lambda \Phi(t_k)).
    \end{split}
\end{equation}
Multiplying both sides by $e^{\lambda t_k}$ (noting that $e^{-\lambda h} = \gamma$):
\begin{equation}
    \gamma \Phi(s_{k+1}) \approx \Phi(s_k) + h (\dot{\Phi}(t_k) - \lambda \Phi(t_k)).
\end{equation}
Rearranging terms yields:
\begin{equation}
    \gamma \Phi(s_{k+1}) - \Phi(s_k) \approx h (\dot{\Phi}(t_k) - \lambda \Phi(t_k)).
\end{equation}
\textbf{Conclusion:} The discrete potential-based reward shaping term $\gamma \Phi_{next} - \Phi_{curr}$ is exactly the update step of the Forward Euler method applied to the differential equation of the discounted potential. This proves that our method is structurally consistent with the underlying continuous-time physics of the task.

\subsection{Policy Invariance under GRM (Proof 5)}
\label{app:proof5}

In this subsection, we prove that adding the shaping term derived above preserves the optimal policy. We show this by demonstrating that the cumulative shaped reward telescopes to a boundary term that is independent of the policy's actions.

Let the shaped reward be $r_{GRM} = r_{env} + F$, where $F(s_t, s_{t+1}) = \gamma\Phi(s_{t+1}) - \Phi(s_t)$. The shaping component of the Q-function, $S^{\pi}(s,a)$, is the expected sum of discounted shaping rewards:
\begin{equation}
    S^{\pi}(s,a) = \mathbb{E}_{\pi} \left[ \sum_{t=0}^{\infty} \gamma^t (\gamma\Phi(s_{t+1}) - \Phi(s_t)) \right].
\end{equation}
We analyze the finite horizon sum $G^{F}_T$:
\begin{equation}
\begin{split}
    G^{F}_T &= \sum_{t=0}^{T-1} \gamma^{t+1} \Phi(s_{t+1}) - \sum_{t=0}^{T-1} \gamma^t \Phi(s_t) \\
    &= \left[ \sum_{k=1}^{T} \gamma^k \Phi(s_k) \right] - \left[ \Phi(s_0) + \sum_{t=1}^{T-1} \gamma^t \Phi(s_t) \right].
\end{split}
\end{equation}
This is a telescoping sum. The intermediate terms cancel out, leaving only the boundary terms:
\begin{equation}
    G^{F}_T = \gamma^T \Phi(s_T) - \Phi(s_0).
\end{equation}
Assuming $\Phi \in [0, 1]$ and $\gamma < 1$, taking the limit $T \to \infty$ yields $\lim_{T\to\infty} \gamma^T \Phi(s_T) = 0$. Thus:
\begin{equation}
    \sum_{t=0}^{\infty} \gamma^t F(s_t, s_{t+1}) = -\Phi(s_0).
\end{equation}
This matches the integral of the continuous form derived in Proof 4: $\int_{0}^{\infty} \frac{d}{dt}(e^{-\lambda t}\Phi) dt = [e^{-\lambda t}\Phi]_0^\infty = -\Phi(0)$.
Since the shaping term evaluates to $-\Phi(s)$ (where $s$ is the state at $t=0$) regardless of the future trajectory, the Q-values are shifted by a state-dependent constant:
\begin{equation}
    Q_{GRM}^{\pi}(s, a) = Q_{env}^{\pi}(s, a) - \Phi(s).
\end{equation}
This shift preserves the ordering of actions:
\begin{gather}
    Q_{GRM}^{\pi}(s, a_1) \ge Q_{GRM}^{\pi}(s, a_2) \nonumber \\
    \Updownarrow \\
    Q_{env}^{\pi}(s, a_1) \ge Q_{env}^{\pi}(s, a_2). \nonumber
\end{gather}
Therefore, the optimal policy remains: $\pi^*_{GRM} = \pi^*_{env}$.

\section{Details of GRM Training Data}
\label{app:data}

\begin{figure*}[t]
    \centering
    \includegraphics[width=0.98\linewidth]{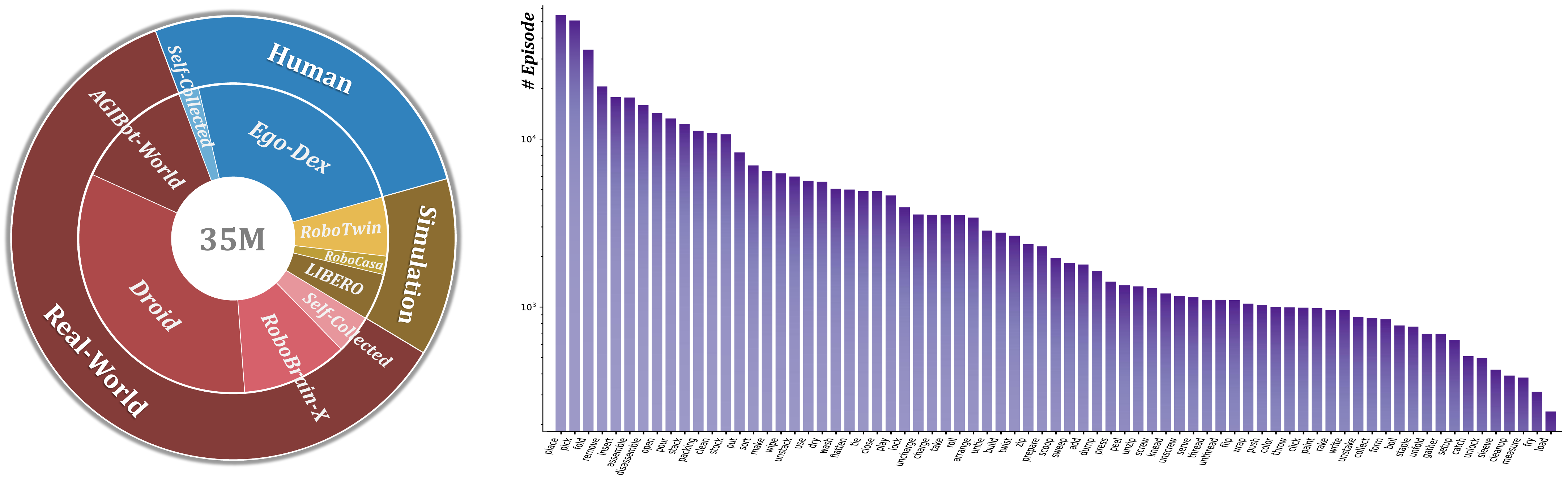}
    \caption{\textbf{Overview of GRM training data.} \textbf{(Left)} The hierarchical composition of our 35M-sample training corpus. The dataset is derived from episodes spanning Real-World Robotics, Simulation, and Human-Centric domains, and is further expanded via multi-view augmentation. \textbf{(Right)} The long-tail distribution of task categories sorted by episode count (log scale). The dataset covers a broad spectrum of manipulation skills, ranging from atomic primitives (\textit{e.g.,} pick, push) to complex, multi-stage horizons (\textit{e.g.,} assemble, fold).}
    \label{fig:reward_profile}
\end{figure*}

In this section, we present a comprehensive breakdown of the training data used to construct our General Reward Model (GRM). To ensure the model possesses robust generalizability across diverse embodiments, environments, and semantic tasks, we curated a massive corpus comprising over 35 million training samples derived from approximately 3,400 hours of raw video footage. This dataset aggregates diverse sources spanning real-world robotics, high-fidelity simulation, and human-centric interactions. We begin by categorizing the constituent datasets (Section~\ref{app:data_source}), followed by an analysis of the embodiment diversity and task statistics (Section~\ref{app:data_stats}). Finally, we detail the rigorous data processing pipeline, specifically our \textit{Stratified Relative Progress Sampling} strategy (Section~\ref{sec:sampling_strategy}), which transforms raw trajectories into balanced and high-quality supervision signals for reward learning.

\subsection{Data Sources}
\label{app:data_source}
Our training data is composed of the following established datasets and self-collected supplements:

\subsubsection{Real-World Datasets}
\begin{itemize}
    \item \textbf{AGIBot-World~\cite{bu2025agibot}:} A large-scale bimanual manipulation dataset collected on the AGIBot-A2D humanoid platform. It provides approximately 3.4M samples focusing on high-dimensional, dual-arm coordination tasks and contact-rich interactions, serving as a critical source for humanoid embodiment generalization.
    \item \textbf{DROID~\cite{khazatsky2024droid}:} A distributed robot interaction dataset collected across multiple institutions. It contributes 8.98M samples featuring the Franka Emika Panda robot. DROID is characterized by its extreme diversity in background scenes, lighting conditions, and object instances, providing robust priors for visual perception.
    \item \textbf{RoboBrain-X~\cite{FlagOpen_RoboBrainX0}:} A comprehensive skill-learning dataset covering a wide range of everyday manipulation skills. We utilize subsets totaling $\sim$3.0M samples, collected on agile platforms (\textit{e.g.,} Agilex Piper, Galaxea R1, AGIBot-A2D). This data enriches the GRM with fine-grained motion primitives.
    \item \textbf{Self-Collected (Real):} To bridge the domain gap between open datasets and our specific experimental setups, we collected an additional 1.1M samples with the RoboOS~\cite{tan2025roboos,tan2025roboosnext}. These include specific long-horizon tasks (\textit{e.g.,} folding, assembly) and corner cases to enhance model robustness.
\end{itemize}
Collectively, these real-world datasets ground our GRM in physical reality, effectively mitigating the ``sim-to-real'' gap often observed in reward modeling. The combination of DROID's extreme visual diversity with the complex morphologies present in AGIBot-World and RoboBrain-X ensures that the model learns robust, embodiment-invariant representations. This allows the GRM to remain stable across varying lighting conditions, textures, and kinematic structures, which is essential for reliable deployment in unstructured environments.

\subsubsection{Simulation Datasets}
\begin{itemize}
    \item \textbf{LIBERO~\cite{liu2023libero}:} A benchmark originally designed for lifelong robot learning. We incorporate 1.33M samples from its task suites (Spatial, Object, Goal, 100), which provide procedurally generated tasks with language instructions, aiding the model in aligning visual progress with semantic goals.
    \item \textbf{RoboCasa~\cite{nasiriany2024robocasa}:} A large-scale simulation framework focused on everyday kitchen activities. It leverages Generative AI to create diverse assets and layouts. We use 0.52M samples to capture realistic object interactions and complex scene semantics in a controlled environment.
    \item \textbf{RoboTwin~\cite{chen2025robotwin}:} A high-fidelity digital twin dataset designed for bimanual manipulation. We utilize a paired dual-view configuration (yielding 1.68M samples from 839k trajectories), which helps the model learn geometry-aware representations crucial for depth disambiguation.
\end{itemize}
While real-world data offers physical fidelity, these simulation environments provide a controlled testbed for learning precise semantic and geometric alignments. The procedural generation inherent in LIBERO and RoboCasa exposes the model to a vast combinatorial space of task instructions and object layouts, fostering strong instruction-following capabilities. Furthermore, the clean, occlusion-free labels and paired views from RoboTwin allow the GRM to learn fine-grained geometric correspondences that are often noisy or unavailable in real-world data.

\subsubsection{Human-Centric Datasets}
\begin{itemize}
    \item \textbf{EgoDex~\cite{hoque2025egodex}:} A large-scale egocentric video dataset capturing dexterous human hand-object interactions. With 6.61M samples, this dataset provides strong priors for understanding hand-object affordances and manipulation logic independent of robot morphology.
    \item \textbf{Self-Collected (Human):} We supplemented the human data with 0.57M samples of domain-specific demonstrations, ensuring coverage of tasks analogous to our robot evaluation protocols.
\end{itemize}
Integrating human video data is pivotal for scaling general manipulation intelligence beyond the limits of available robot demonstrations. By leveraging the massive scale of EgoDex, our GRM acquires universal object affordance priors, learning ``how'' objects should be manipulated regardless of the specific actuator. This cross-embodiment transfer is critical for evaluating progress in novel tasks where robot-specific data may be scarce, enabling the reward model to generalize purely based on the observed state changes of the objects themselves.

\subsection{Statistics and Embodiment Diversity}
\label{app:data_stats}

\begin{table}[t]
    \centering
    \caption{\textbf{Statistics of Raw Data Sources.} The table lists the \textit{raw} frame counts (in thousands, `k') before augmentation. The final training set is expanded to \textbf{35M} via multi-view expansion and augmentation strategies.}
    \label{tab:data_source_stats}
    \resizebox{0.95\linewidth}{!}{%
    \begin{tabular}{l|l|r}
        \toprule
        \textbf{Domain} & \textbf{Dataset Source} & \textbf{Raw Samples} \\
        \midrule
        \multirow{5}{*}{Real-World} 
        & AGIBot-World~\cite{bu2025agibot} & 3,400k \\
        & DROID~\cite{khazatsky2024droid} & 8,983k \\
        & RoboBrain-X~\cite{FlagOpen_RoboBrainX0} & 3,025k \\
        & Self-Collected (Real) & 1,107k \\
        \cmidrule{2-3}
        & \textit{\textbf{Subtotal}} & \textit{16,515k} \\
        \midrule
        \multirow{4}{*}{Simulation} 
        & LIBERO~\cite{liu2023libero} & 1,330k \\
        & RoboTwin~\cite{chen2025robotwin} & 1,678k \\
        & RoboCasa~\cite{nasiriany2024robocasa} & 523k \\
        \cmidrule{2-3}
        & \textit{\textbf{Subtotal}} & \textit{3,531k} \\
        \midrule
        \multirow{3}{*}{Human} 
        & EgoDex~\cite{hoque2025egodex} & 6,610k \\
        & Self-Collected (Human) & 574k \\
        \cmidrule{2-3}
        & \textit{\textbf{Subtotal}} & \textit{7,184k} \\
        \midrule
        \textbf{Total} & \textbf{Raw Corpus} & \textbf{27,230k} \\
        \bottomrule
    \end{tabular}
    }
\end{table}

\begin{table}[t]
    \centering
    \caption{\textbf{Distribution by Robot Embodiment.} Our dataset spans diverse robot hardware, from single-arm industrial robots to bimanual humanoids, ensuring strong physical generalization.}
    \label{tab:embodiment_stats}
    \resizebox{0.98\linewidth}{!}{%
    \begin{tabular}{l|l|r}
        \toprule
        \textbf{Robot Embodiment} & \textbf{Primary Data Sources} & \textbf{Samples} \\
        \midrule
        Franka Emika Panda & DROID, LIBERO, RoboCasa & 8,983k \\
        AGIBot-A2D & AGIBot-World, RoboBrain-X & 3,400k \\
        Agilex Piper & RoboBrain-X, Self-Collected & 3,552k \\
        ARX-X5 & RoboTwin & 882k \\
        Galaxea & RoboBrain-X, Self-Collected & 695k \\
        UR5 & Self-Collected & 433k \\
        \bottomrule
    \end{tabular}
    }
\end{table}

Our compiled dataset represents one of the most comprehensive collections for robotic manipulation to date. As detailed in~\Cref{tab:data_source_stats}, the final training corpus of 35 million samples is strategically balanced to maximize generalization. Real-world interaction data constitutes the majority ($\sim$60\%) to ensure the model is grounded in physical reality, while high-fidelity simulation ($\sim$13\%) and large-scale human videos ($\sim$26\%) provide necessary semantic breadth and affordance priors that are difficult to obtain from robots.

\textbf{Embodiment Agnosticism.} A core design principle of our GRM is robustness to morphological variations. As illustrated in~\Cref{tab:embodiment_stats}, the dataset covers a wide spectrum of robot embodiments, preventing the model from overfitting to specific kinematics or camera calibrations. By training on this heterogeneous mixture, the GRM learns to focus on the \textit{state changes of the objects} rather than the robot's motion, enabling zero-shot transfer to unseen robot configurations.

\textbf{Task and Instruction Diversity.}~\Cref{fig:reward_profile} (Right) highlights the semantic diversity of the dataset. The task distribution follows a natural long-tail pattern, spanning from high-frequency atomic primitives (\textit{e.g.,} \texttt{pick}, \texttt{place}, \texttt{push}) that build a solid generalist foundation, to complex, multi-stage horizons (\textit{e.g.,} \texttt{fold}, \texttt{assemble}, \texttt{pour}) that challenge the model's ability to track long-term progress. To further enhance semantic robustness, we employed Gemini 2.5 Pro~\cite{comanici2025gemini} to re-annotate task instructions, thereby introducing linguistic diversity and reducing overfitting to rigid template prompts.

\subsection{Sampling Strategy and Data Balancing}
\label{sec:sampling_strategy}

To train the General Reward Model (GRM) effectively, it is crucial to generate training pairs that cover the full spectrum of task progress dynamics. Naively sampling random pairs from a trajectory typically results in a long-tail distribution dominated by small, positive progress steps, leading to model bias. To address this, we use the Stratified Relative Progress Sampling strategy combined with an aggressive augmentation pipeline, as described in main paper. More details are as follow:

\subsubsection{Hop-based Relative Progress Formulation}
Consistent with the main paper, we define the ground-truth relative progress label $\mathcal{H}(s_p, s_q)$ using a \textit{relative-relative} formulation. Let a trajectory consist of $M$ steps $\{s_0, \dots, s_M\}$ with linear global progress $\Phi(s_i) = i/M$. Without loss of generality, we normalize the global progress such that the initial state has $\Phi(s_0) = 0$ and the goal state has $\Phi(s_M) = 1$. For any training pair $(s_p, s_q)$:

\begin{equation}
\mathcal{H}(s_p, s_q) = 
\begin{cases} 
  \dfrac{\Phi(s_q) - \Phi(s_p)}{1 - \Phi(s_p)} & \text{if } q \geq p \textsc{ (progress)} \\
  \\
  \dfrac{\Phi(s_q) - \Phi(s_p)}{\Phi(s_p)} & \text{if } q < p \textsc{ (regress)}.
\end{cases}
\end{equation}

\begin{figure*}[t]
    \centering
    \includegraphics[width=0.98\linewidth]{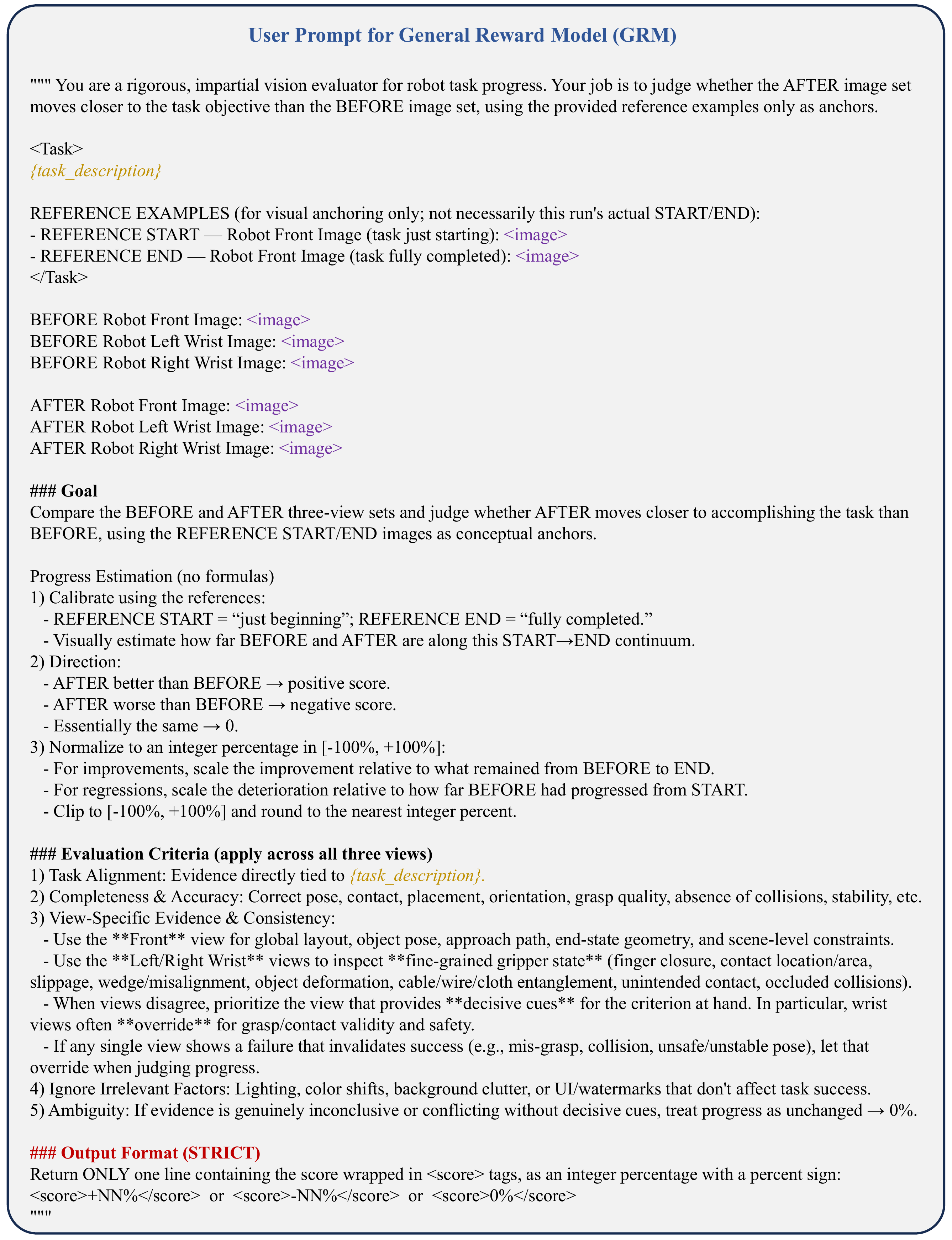}
    \caption{User Prompt for General Reward Model (GRM).}
    \label{fig:prompt}
\end{figure*}

This formula ensures that forward progress is normalized by the \textit{remaining distance} ($1 - \Phi(s_p)$), making late-stage steps statistically significant, while regression is normalized by the \textit{accumulated distance} ($\Phi(s_p)$). The continuous output $\mathcal{H} \in [-1, 1]$ is quantized into integer percentage tokens for VLM training. See~\Cref{fig:prompt} for the whole prompt.

\subsubsection{Hierarchical Score-Gap Stratified Sampling}
To prevent the model from overfitting to specific step sizes, we implement a two-stage sampling mechanism:

\begin{itemize}
    \item \textbf{Score Balancing:} We discretize the progress score range $[-100\%, +100\%]$ into $N_{score}$ uniform bins (\textit{e.g.,} $N_{score}=25$). We generate a large candidate pool of random pairs and fill these bins to enforce a uniform distribution of reward values. This ensures the model encounters rare events, such as significant regression (errors) or large forward jumps, as frequently as common small steps.
    \item \textbf{Temporal Gap Diversification:} Within each score bin, we further categorize samples based on their temporal distance $\Delta t = |t_{post} - t_{pre}|$ into $N_{gap}$ bins (\textit{e.g.,} $N_{gap}=4$). This method explicitly forces the model to distinguish between \textit{fast progress} (large score change in short time) and \textit{slow progress} (large score change over long time), effectively decoupling visual state change from the temporal duration of trajectories.
\end{itemize}

\subsubsection{Zero-Hop Anchoring}
Standard comparative ranking often struggles with static or near-static pairs, leading to hallucinated progress values. To mitigate this, we explicitly inject a fixed ratio $\alpha$ (\textit{e.g.,} $\alpha=5\%$) of \textbf{\textit{Zero-Score Samples}}. These samples are constructed by selecting pairs $(t, t+\delta)$ where the temporal delta $\delta$ is negligible (randomly sampled within $\pm 10\%$ of the local window). These pairs are labeled with a strict score of $0\%$. This creates a ``semantic anchor'' for the model, teaching it that visual similarity implies zero progress, thereby reducing false positives in stable robotic states.

\subsubsection{Data Augmentation and Robustness}
Finally, to bridge the gap between the raw data count and our effective training scale, and to ensure robustness against perceptual failures, we employ an aggressive data augmentation pipeline:

\begin{itemize}
    \item \textbf{Multi-View Expansion:} We maximize the utility of multi-camera setups (\textit{e.g.,} permuting available wrist and third-person views) by treating distinct viewpoints as complementary training signals. This strategy expands the dataset size from 27.2M raw samples to 35M training samples, improving the model's geometric consistency.
    \item \textbf{Perceptual Robustness Training:} To prevent the model from over-relying on any single input modality, we introduce \textit{Random Viewpoint Dropout} (randomly masking specific camera feeds) and \textit{Context Dropout} (randomly masking $s_{init}$ or $s_{goal}$ tokens) during training. This forces the GRM to infer progress from partial or occluded observations, significantly enhancing resilience in real-world deployments.
\end{itemize}

\section{Experiment Details}
\label{app:exp}

\begin{figure}[t]
    \centering
    \includegraphics[width=0.98\linewidth]{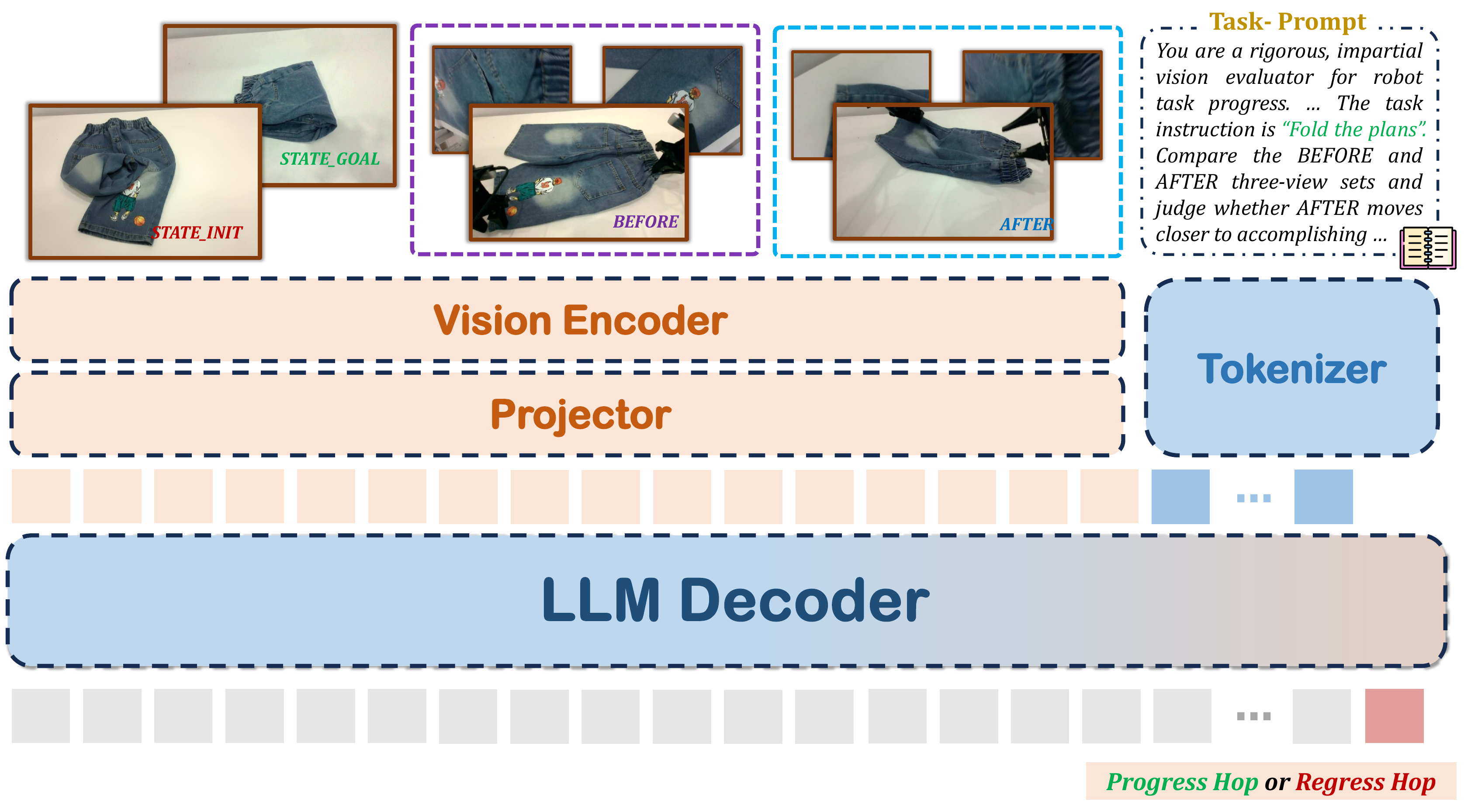}
    \caption{\textbf{Overview of GRM model structure.} The GRM is built upon the Qwen2.5-VL architecture. It processes a multimodal interleaved input sequence consisting of task text instructions and multi-view images: the initial state ($s_{init}$), the goal state ($s_{goal}$), and the paired ``BEFORE'' ($s_{pre}$) and ``AFTER'' ($s_{post}$) observation sets. The visual signals are processed by a shared Vision Encoder and Projector, then fed into the LLM Decoder, which autoregressively predicts a quantized relative progress token (\textit{e.g.,} \texttt{<score>+15\%</score>}) or a regress token (\textit{e.g.,} \texttt{<score>-4\%</score>}).}
    \label{fig:dp_model}
\end{figure}

\begin{table}[t]
\centering
\caption{GRM Training Hyperparameters.}
\label{tab:train_hyperparams}
\resizebox{0.90\linewidth}{!}{%
\begin{tabular}{l|cc}
\toprule
\textbf{Hyperparameter} & \textbf{GRM-3B} & \textbf{GRM-8B} \\
\midrule
Global Batch Size & 512 & 256 \\
LR: $\{\psi_v^{\text{ViT}}, \phi_v^{\text{LLM}}\}$ & $\{5\text{e}^{-6}, 1\text{e}^{-5}\}$ & $\{5\text{e}^{-6}, 1\text{e}^{-5}\}$ \\
LR Scheduler & Cosine Decay & Cosine Decay \\
Warmup Ratio & 0.03 & 0.03 \\
Optimizer & AdamW & AdamW \\
Weight Decay & 0.1 & 0.1 \\
Max Sequence Length & 8192 & 8192 \\
Tensor Parallelism (TP) & 1 & 2 \\
Pipeline Parallelism (PP) & 1 & 2 \\
GPU nodes & 16$\times$8 H100 & 16$\times$8 H100 \\
\midrule
Training Duration & $\sim$ 8 Days & $\sim$ 14 Days \\
\bottomrule
\end{tabular}
}
\end{table}

\subsection{GRM Training and Evaluation}
\label{app:grm-exp}

\textbf{Model Details.}
Our GRM leverages the RoboBrain 2.0~\cite{team2025robobrain} architecture, as shown in~\Cref{fig:dp_model}, exploring two parameter scales: a lightweight 3B variant for efficient inference and a powerful 8B variant for maximum reasoning capability. 
For inputs, the model accepts a multimodal prompt sequence. Visual inputs include the Initial State $I_{init}$, Goal State $I_{goal}$, and two sets of multi-view observations for the transition being evaluated: $S_{pre} = \{I_{pre}^{v}\}_{v=1}^V$ and $S_{post} = \{I_{post}^{v}\}_{v=1}^V$, where $V$ is the number of available camera views (\textit{e.g.,} one front, two wrists). Textual inputs include the system prompt defining the rigorous evaluator persona and the specific task instruction $d_{task}$.
For outputs, the model outputs a discrete token representing the relative progress hop $\mathcal{H}(S_{pre}, S_{post}) \in [-1, 1]$.

\textbf{Training Infrastructure.}
We conducted large-scale training on a high-performance cluster consisting of 128 $\times$ NVIDIA H100 (80GB) GPUs. The training framework utilizes the Megatron-LM architecture~\cite{shoeybi2019megatron} for efficient distributed training. We employed Tensor Parallelism (TP) and Pipeline Parallelism (PP) to maximize throughput. Detailed hyperparameters are provided in~\Cref{tab:train_hyperparams}. The 3B model was trained for approximately 8 days, while the 8B model required 14 days to converge on the 35M-sample dataset.

\textbf{Evaluation Protocols.}
We evaluate the GRM on two distinct tasks to assess both its fine-grained temporal understanding and its high-level task success judgment via vLLM engine~\cite{kwon2023efficient}.
For \textit{Video Frame Rank-Correlation}, we use Value-Order Consistency (VOC)~\cite{ma2024gvl} as main metric, which assesses whether the reward model correctly orders states based on temporal progress. We measure VOC on unseen test data, where details are as follow:
\begin{itemize}
    \item \textbf{Data Selection:} For each dataset (\textit{e.g.,} AgiBot-World, DROID), we randomly sample 10 trajectories per subclass in each dataset from the hold-out validation set.
    \item \textbf{Metric:} Given two frames $t_A$ and $t_B$ where $t_A < t_B$, a correct prediction requires $R(s_{t_A}) < R(s_{t_B})$. The VOC score is the correlation coefficient $[-1, 1]$.
    \item \textbf{Baselines:} We compare with VLAC~\cite{zhai2025vlac} and GVL~\cite{ma2024gvl} using their official codebase\footnote{\url{https://github.com/InternRobotics/VLAC}} or recommended prompts.
\end{itemize}
For \textit{Task Completion Judgment}, to verify if the GRM can serve as a reliable success detector, we collected 60 real-world rollouts for each of three challenging tasks: \textit{Stacking Blocks}, \textit{Folding T-shirt / Pants}, and \textit{Clearing Desktop}. These rollouts are uniformly distributed into 20 Successful (SE), 20 Partially Successful (PSE), and 20 Failed (FE) episodes.
We define the automated judgment logic following SARM~\cite{chen2025sarm}. Let $P_t \in [0, 1]$ be the accumulated global progress at step $t$ predicted by the GRM. The trajectory classification label is determined by:
\begin{equation}
\label{eq:judgment_logic}
\text{Label} = 
\begin{cases} 
\text{SE}, & \begin{aligned}[t]
& \text{if } P_{\text{final}} > 0.8 \ \land \frac{3}{T} \sum_{t=2T/3}^{T} P_t > 0.6, 
\end{aligned} \\
\text{PSE}, & \text{if } \frac{1}{T} \sum_{t=1}^{T} P_t \ge \xi, \\
\text{FE}, & \text{otherwise.}
\end{cases}
\end{equation}
where $\xi=0.4$ is a threshold for partial progress.

\subsection{Simulation Experiments}
\label{app:sim-exp}

This section details the simulation experiments: the benchmark and two different ``RL algorithm + VLA model'' settings we employ to validate our methods.

\subsubsection{Benchmark}
We evaluate our framework on the LIBERO-Goal suite~\cite{liu2023libero}, a challenging subset of the LIBERO benchmark consisting of 10 distinct long-horizon manipulation tasks. These tasks require the agent to reason about specific goal configurations and perform precise object interactions, serving as a rigorous testbed for progress-based reward modeling.

\subsubsection{Setting 1: PPO + OpenVLA-OFT}

In this setting, we combine PPO (Proximal Policy Optimization)~\cite{ppo}, a stable and sample-efficient on-policy RL algorithm, with OpenVLA-OFT~\cite{openvla-oft}, an optimized vision-language-action (VLA) model, to build the reinforcement learning pipeline, implementing it via the RLinf codebase~\cite{zang2025rlinf}. OpenVLA-OFT acts as the base policy model, refined from the OpenVLA backbone through an Optimized Fine-Tuning (OFT) recipe that integrates parallel decoding with action chunking (for high-throughput action generation), continuous action representation (avoiding lossy discretization), and an L1 regression objective (simplifying training while preserving precision). It maps multimodal inputs to continuous robot actions in one forward pass. For LIBERO-Goal’s long-horizon tasks, we set an action chunk size of 8 to capture temporal dependencies and reduce compounding errors.

PPO optimizes the OpenVLA-OFT policy by confining updates to a trust region, using a clipped surrogate objective to balance exploration and exploitation. Its core optimization objective is defined as:
\begin{equation}
\small
J^{\text{PPO}}(\theta) = \mathbb{E}_{t} \left[ \min \left( \rho_t(\theta) \hat{A}_t, \text{clip}\left(\rho_t(\theta), 1-\epsilon, 1+\epsilon\right) \hat{A}_t \right) \right].
\end{equation}
Here, $\rho_t(\theta) = \frac{\pi_{\theta}(a_t | o_t)}{\pi_{\theta_{\text{old}}}(a_t | o_t)}$ is the ratio between the current policy $\pi_\theta$ and the fixed rollout policy $\pi_{\theta_{\text{old}}}$, $\hat{A}_t$ is the estimated advantage at timestep $t$, and $\epsilon$ is the clipping parameter that prevents excessive updates. We follow RLinf-VLA~\cite{zang2025rlinf}’s PPO best practices: using action-level value estimation (superior for chunked policies) and partial reset (resetting sub-environments post-completion to boost sample efficiency).

\subsubsection{Setting 2: ReinFlow + $\pi_0$}
\begin{figure}
    \centering
    \includegraphics[width=1\linewidth]{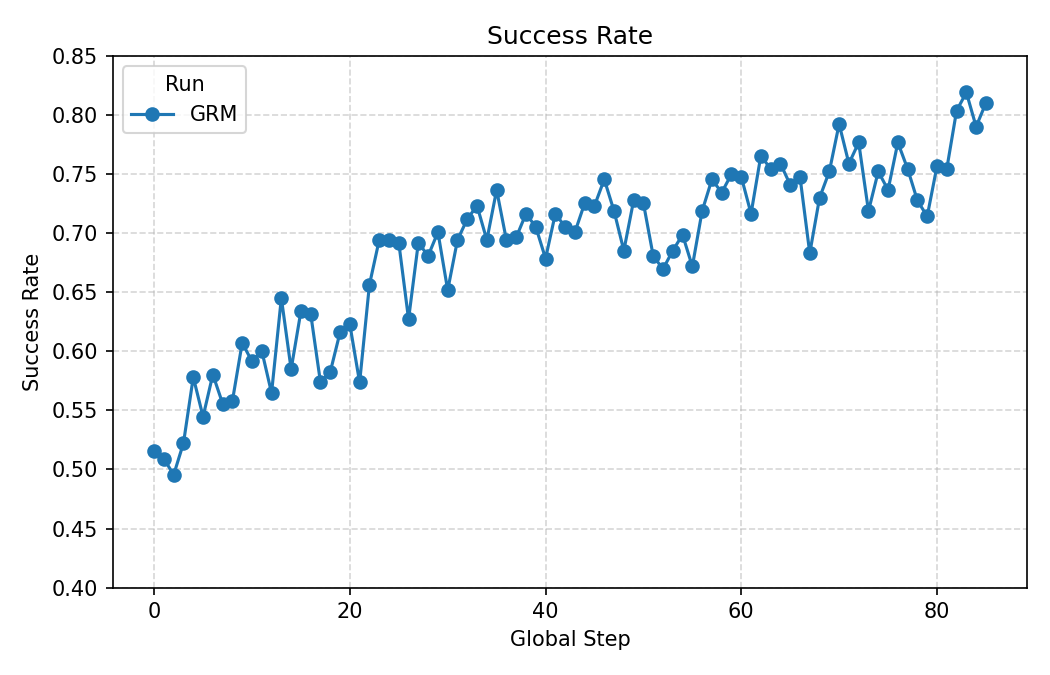}
    \caption{\textbf{Training Curve on LIBERO-Goal.} The success rate of our ReinFlow agent fine-tuned with GRM-based reward shaping. The agent demonstrates stable convergence and achieves an average success rate of over 80\% across the benchmark tasks.}    
    \label{fig:success_rate_libero}
\end{figure}

In this setting, we combine {ReinFlow}~\cite{reinflow}, a flow-matching based reinforcement learning algorithm, and \textbf{$\pi_0$}~\cite{pi0}, a flow-matching based vision-language-action model to build the reinforcement learning pipeline and implement it using the RLinf~\cite{zang2025rlinf} codebase\footnote{\url{https://github.com/RLinf/RLinf}}. The training is conducted on a compute node equipped with 8 NVIDIA H100 GPUs. The total training time is approximately 50 hours, achieving an average success rate of 81\% across the LIBERO-Goal tasks (Figure \ref{fig:success_rate_libero}).
ReinFlow facilitates stable fine-tuning by injecting learnable noise into the flow matching policy's deterministic path, converting it into a discrete-time Markov process. Specifically, the denoising process is modified as follows:
\begin{equation}
    a^{k+1} \sim \mathcal{N}\left( a^k + v_\theta(t_k, a^k, o)\Delta t_k, \, \sigma_{\theta'}^2(t_k, a^k, o) \right)
\end{equation}
where $v_\theta$ is the velocity network and $\sigma_{\theta'}$ is a learnable noise network. This formulation allows for an exact computation of the transition probability:
\begin{equation}
\begin{split}
    \pi^\theta(a^{k+1}|a^k, o) &= \ln \mathcal{N}\left( a^{k+1} \bigg| a^k \right. \\
    &\quad \left. + v_\theta(t_k, a^k, o)\Delta t_k, \, \sigma_{\theta'}^2(t_k, a^k, o) \right)
\end{split}
\end{equation}
The policy is then optimized using a PPO-style objective that jointly updates the velocity and noise networks: 
\begin{equation}
\begin{split}
    \theta, \theta' =& \arg\min_{\theta, \theta'} \sum_{i=1}^{B} \left[ -A^{\bar{\theta}_{old}}(o_i, a_i) \sum_{k=0}^{K-1} \ln \pi^{\bar{\theta}}(a_i^{k+1}|a_i^{k}, o_i) \right. \\
     & +\left. \alpha \cdot \mathcal{R}(a_i, o_i) \right]
\end{split}
\end{equation}
where $A$ is the advantage function, and $\mathcal{R}$ is a regularization term (\textit{e.g.,} entropy).
Furthermore, detailed hyperparameters for the ReinFlow training are provided in Table~\ref{tab:sim_hyperparams}.

\begin{table}[h]
    \centering
    \small
    \caption{Hyperparameters for LIBERO-Goal Experiments.}
    \label{tab:sim_hyperparams}
    \begin{tabular}{l|c}
        \toprule
        \textbf{Hyperparameter} & \textbf{Value} \\
        \midrule
        Algorithm & ReinFlow \\
        Consistency Sensitivity $\alpha$ & 20 \\
        Discount Factor $\gamma$ & 0.99 \\
        Batch Size & 1792 \\
        Learning Rate & 5e-6 \\
        Compute Resources & 8 $\times$ H100 \\
        \bottomrule
    \end{tabular}
\end{table}

\subsection{Real-World Experiments}
\label{app:real-exp}

This section details the real-world robotic experiments: the algorithm we use and 8 representative tasks we conduct.

\subsubsection{Algorithm}

We adopt the Consistency-based Reinforced Fine-Tuning (ConRFT) algorithm~\cite{conrft} in our real-world experiments. ConRFT is an Cal-QL~\cite{calql} algorithm variant designed as an offline-to-online reinforcement learning (RL) method rooted in Q-learning for fine-tuning pre-trained Vision-Language-Action (VLA) models. ConRFT extends Cal-QL’s framework by adding a consistency policy and a behavior cloning loss, enabling efficient and safe adaptation for real-world robotic manipulation tasks. Key components are as follows:

\textbf{Algorithm Overview.} ConRFT operates in two sequential phases: offline initialization and online adaptation. It maintains two data buffers: a {demonstration buffer} ($\mathcal{D}$) and a {replay buffer} ($\mathcal{R}$). The offline phase leverages a small set of human demonstrations (20-30 trajectories) stored in $\mathcal{D}$ to initialize a policy via behavior cloning (BC) and calibrated Q-learning (Cal-QL)~\cite{calql}, ensuring stable and safe initial behavior without requiring real-world exploration. The online phase then refines this policy through interaction with the physical world, storing transitions in $\mathcal{R}$. A human-in-the-loop (HIL) component intervenes to correct unsafe actions during online deployment, with these corrected trajectories added back to $\mathcal{D}$ to guide subsequent policy updates. Symmetric sampling from the combined dataset $\mathcal{D} \cup \mathcal{R}$ is employed to mitigate distribution shift between the offline and online stages.
  
\textbf{Policy model}. The policy model is built on the Octo-Small model~\cite{octo}, a lightweight VLA backbone selected for its balance of performance and inference efficiency. The model is trained with visual encoders and transformer backbone and its original action head is replaced with a consistency policy~\cite{consistency}, a diffusion-based module that maps Gaussian-sampled random actions to expert-like behaviors, conditioned on the observation embeddings.

\textbf{Offline Stage.} The policy is initialized exclusively using data from the demonstration buffer $\mathcal{D}$, with a diffusion-based consistency policy (parameterized by $\psi$) serving as the action head. The unified training objective integrates BC and Q-learning to balance expert imitation and reward alignment, defined as 
\begin{equation}
\mathcal{L}_\pi^{\text{offline}}(\psi) = \beta \mathcal{L}_\pi^{\text{BC}} + \eta \mathcal{L}_\pi^{\text{Q}},
\end{equation}
where $\beta$ and $\eta$ are hyperparameters that balance the two loss terms. 
The BC loss minimizes the Euclidean distance between actions generated by the consistency policy and expert demonstrations from $\mathcal{D}$, formulated as
\begin{equation}
\mathcal{L}_\pi^{\text{BC}} = 
\mathbb{E}_{\substack{(s,a) \sim \mathcal{D} \\ m \sim U[1,M-1]}} 
\Bigg[ \| f_\psi(a + k_m z, k_m \mid E_\phi(s)) - a \|_2 \Bigg],
\end{equation}
in which $(s,a) \sim \mathcal{D}$ denotes sampling state-action pairs from the demonstration buffer, $m \sim U[1,M-1]$ indicates random sampling of a diffusion sub-interval, $k_m$ is the diffusion step corresponding to sub-interval $m$, $z \sim \mathcal{N}(0,I)$ is Gaussian noise following a standard normal distribution, $E_\phi(s)$ is the observation embedding output by the frozen Octo-Small backbone, $f_\psi$ is the consistency policy module parameterized by $\psi$, and $\| \cdot \|_2$ denotes the Euclidean distance. The Q loss maximizes the action-value estimates from a learned critic network $Q_\theta$ (where $\theta$ are the critic's parameters), given by 
\begin{equation}
\mathcal{L}_\pi^{\text{Q}} = -\mathbb{E}_{s \sim \mathcal{D}, a \sim \pi_\psi} \left[ Q_\theta(s, a) \right],
\end{equation}
with $a \sim \pi_\psi$ indicating actions sampled from the consistency policy. The critic $Q_\theta$ is updated using the Cal-QL objective: 
\begin{equation}
\begin{split}
\mathcal{L}_Q^{\text{offline}}(\theta) &= \alpha \left( \mathbb{E}_{s \sim \mathcal{D}, a \sim \pi} \left[ \max(Q_\theta(s,a), V^\mu(s)) \right] \right. \\
& \quad - \left. \mathbb{E}_{(s,a) \sim \mathcal{D}} \left[ Q_\theta(s,a) \right] \right) \\
& \quad + \frac{1}{2} \mathbb{E}_{(s,a,s') \sim \mathcal{D}} \left[ \left( Q_\theta(s,a) - \mathcal{B}^\pi \bar{Q}_{\bar{\theta}}(s,a) \right)^2 \right].
\end{split}
\end{equation}
In this formula, $\alpha$ is the conservative penalty hyperparameter, $V^\mu(s)$ is the value of a reference policy $\mu$ that stabilizes estimates for out-of-distribution actions, $(s,a,s') \sim \mathcal{D}$ samples transition triples (state, action, next state) from $\mathcal{D}$, $\mathcal{B}^\pi$ is the Bellman backup operator defined as $\mathcal{B}^\pi \bar{Q}(s,a) = r(s,a) + \gamma \mathbb{E}_{a' \sim \pi} \bar{Q}(s',a')$, and $\bar{Q}_{\bar{\theta}}$ is a delayed target Q-network whose parameters $\bar{\theta}$ are fixed during critic updates to ensure training stability.

\textbf{Online Stage.} The online phase refines the policy using combined data from the demonstration buffer $\mathcal{D}$ which is augmented with corrected trajectories from HIL interventions and the replay buffer $\mathcal{R}$ that stores online interaction transitions. The policy update objective retains the same structure as the offline stage but with adjusted weights to prioritize reward-driven exploration over expert imitation: 
\begin{equation}
\mathcal{L}_\pi^{\text{online}}(\psi) = \beta' \mathcal{L}_\pi^{\text{BC}} + \eta' \mathcal{L}_\pi^{\text{Q}},
\end{equation}
where $\beta'$ is reduced  and $\eta'$ is increased, shifting the focus toward learning from real-world feedback. The BC and Q losses follow the same formulations as in the offline phase, with the only difference being that data sampling is performed from $\mathcal{D} \cup \mathcal{R}$ instead of $\mathcal{D}$ alone, and all symbols retain their previously defined meanings. The critic $Q_\theta$ is updated via a standard Bellman error loss without the conservative penalty (as online data reduces distribution shift and mitigates overestimation risks), given by
\begin{equation}
\mathcal{L}_Q^{\text{online}}(\theta) = \mathbb{E}_{(s,a,s') \sim \mathcal{D} \cup \mathcal{R}} \left[ \left( Q_\theta(s,a) - \mathcal{B}^\pi \bar{Q}_{\bar{\theta}}(s,a) \right)^2 \right],
\end{equation}
ensuring accurate value estimation as the policy adapts to real-world dynamics.

\textbf{Hyperparameters} Detailed hyperparameters for real-world experiments are provided in Table \ref{tab:real_exp_hyper}.

\subsubsection{Real-World Tasks}

In real-world experiments, we selected 8 representative tasks to verify the promotional effect of our reward model on physical robotic reinforcement learning. These tasks cover single-arm tasks, dual-arm tasks, fine-grained tasks, and long-horizon tasks, with detailed descriptions of each task provided as follows.

\textbf{Insert Square:} The end-effector of the robotic arm grasps a square block with four holes. On the table, there is a target board designed for block insertion, which is equipped with four upright pegs corresponding to the four holes on the block. The robot is required to adjust the position and orientation of the block to insert it onto the pegs of the board. This task demands millimeter-level precision tolerance, categorizing it as a single-arm fine-grained task.

\textbf{Pick and Place:} A white gasket and a toy are placed on the table. The robotic arm needs to first grasp the toy and then place it accurately on the gasket. The initial positions of both the gasket and the toy are randomly generated within a predefined range. This task is defined as a single-arm fine-grained long-horizon task.

\textbf{Complete Circuit:} A disconnected circuit is placed on the table, and the right arm of the dual-arm robotic system holds a battery. The robot is required to first insert the battery (held by the right arm) into the battery slot, then use the left arm to toggle the circuit switch, thereby lighting up the bulb in the middle of the circuit. This task is classified as a dual-arm fine-grained long-horizon task.

\textbf{Arrange Flowers:} A vase is placed on the table, with the left and right arms of the robotic system each holding a flower. The robot needs to sequentially insert the two flowers into the vase. This task is categorized as a dual-arm fine-grained long-horizon task.

\textbf{Fold Towel:} A towel is laid on the table, and the robotic arm is required to fold it neatly. This task is defined as a dual-arm fine-grained long-horizon task.

\textbf{Build Blocks:} Three building blocks are placed on the table. The robotic arm needs to stack them sequentially to form a castle-like structure. This task is classified as a dual-arm fine-grained long-horizon task.

\textbf{Cap the Pen:} The dual-arm robot holds a pen in one arm and a pen cap in the other. It needs to slowly adjust the positions and orientations of the two objects to finally fit the cap onto the pen. This task is categorized as a dual-arm fine-grained task.

\textbf{Zip the Bag:} The left arm of the robot holds a bag, while the right arm clamps the zipper of the bag. The robot is required to coordinate the movements of both arms to zip up the bag completely. This task is defined as a dual-arm fine-grained task.

\begin{table}[t]
\centering
\caption{Real-World Experiment Hyperparameters.}
\label{tab:real_exp_hyper}
\resizebox{0.95\linewidth}{!}{%
\begin{tabular}{l|cc}
\toprule
\textbf{Hyperparameter} & \textbf{Symbol} & \textbf{Value} \\
\midrule
Global Batch Size & - & 256 \\
Learning Rate & - & $3 \times 10^{-4}$ \\
Reward Discount Factor & $\gamma$ & $0.98$ \\
Offline BC Loss Weight & $\beta$ & 1.0 \\
Offline Q Loss Weight & $\eta$ & 0.1 \\
Conservative Penalty Coefficient & $\alpha$ & 0.1 \\
Online BC Loss Weight & $\beta'$ & 0.5 \\
Online Q Loss Weight & $\eta'$ & 1.0 \\
Compute Resources & - & 1 $\times$ RTX4090 \\
\bottomrule
\end{tabular}
}
\end{table}

\section{More Qualitative Results}
\label{app:vis}
In this section, we provide additional qualitative visualizations to further substantiate the effectiveness and robustness of our method. We focus on three key aspects: the generalization of GRM across diverse semantic tasks, the temporal robustness of progress estimation under varying sampling intervals, and the trajectory visualization of real-world RL.

\begin{figure*}[t]
    \centering
    \includegraphics[width=0.90\linewidth]{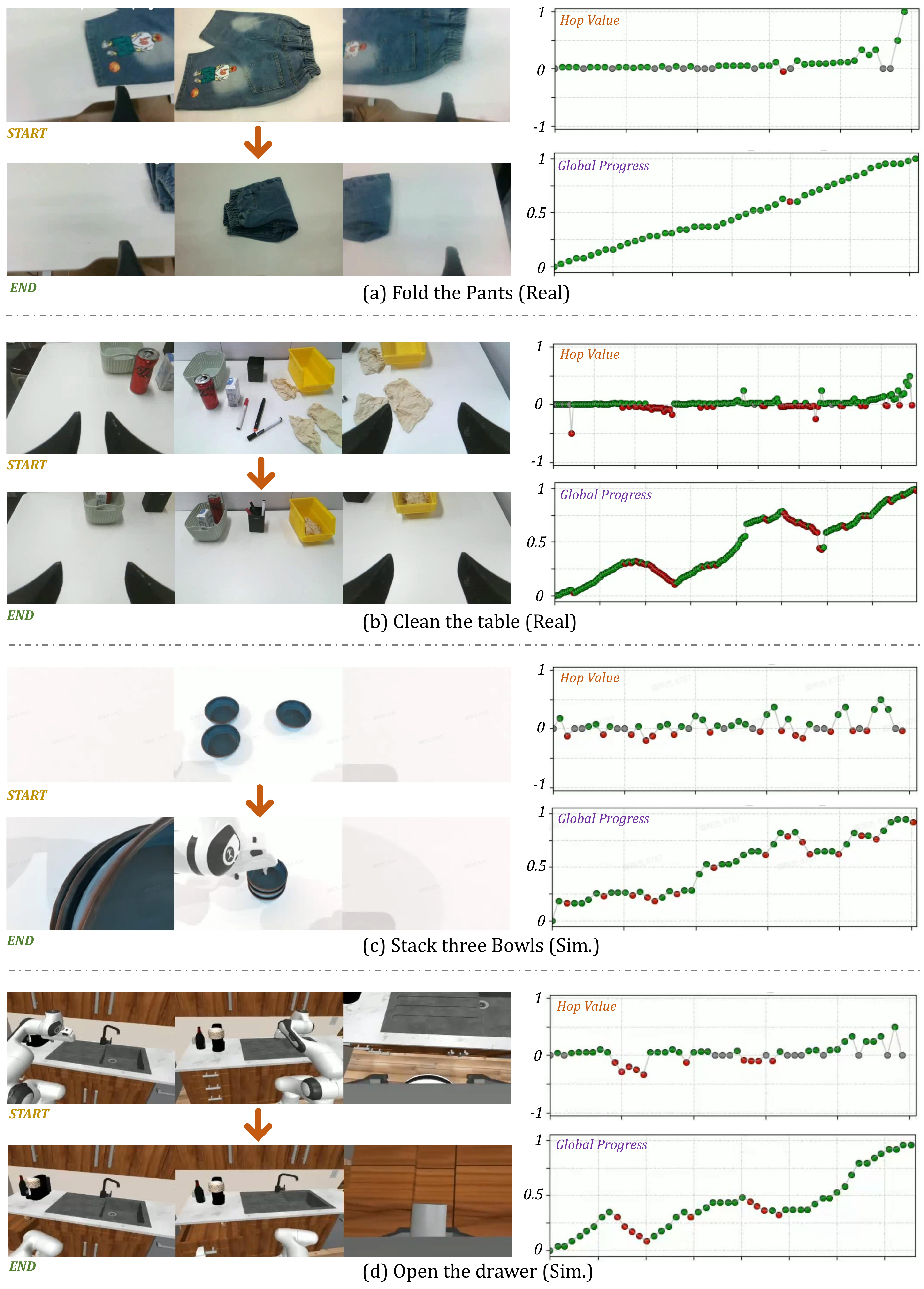}
    \caption{\textbf{GRM Progress Predictions across Diverse Tasks.} We visualize the frame-wise \textit{Hop} (instantaneous change) and accumulated \textit{Progress} predicted by GRM on unseen validation tasks.}
    \label{fig:vis_tasks}
\end{figure*}

\textbf{GRM Predictions on Diverse Tasks.}
Figure~\ref{fig:vis_tasks} illustrates the predicted Hop and Progress curves generated by our GRM across a variety of manipulation tasks. The results demonstrate that our model accurately captures the monotonic increase in progress for successful trajectories while effectively identifying stagnation or regression in failed attempts, validating its semantic generalization capabilities.

\begin{figure*}[t]
    \centering
    \includegraphics[width=0.98\linewidth]{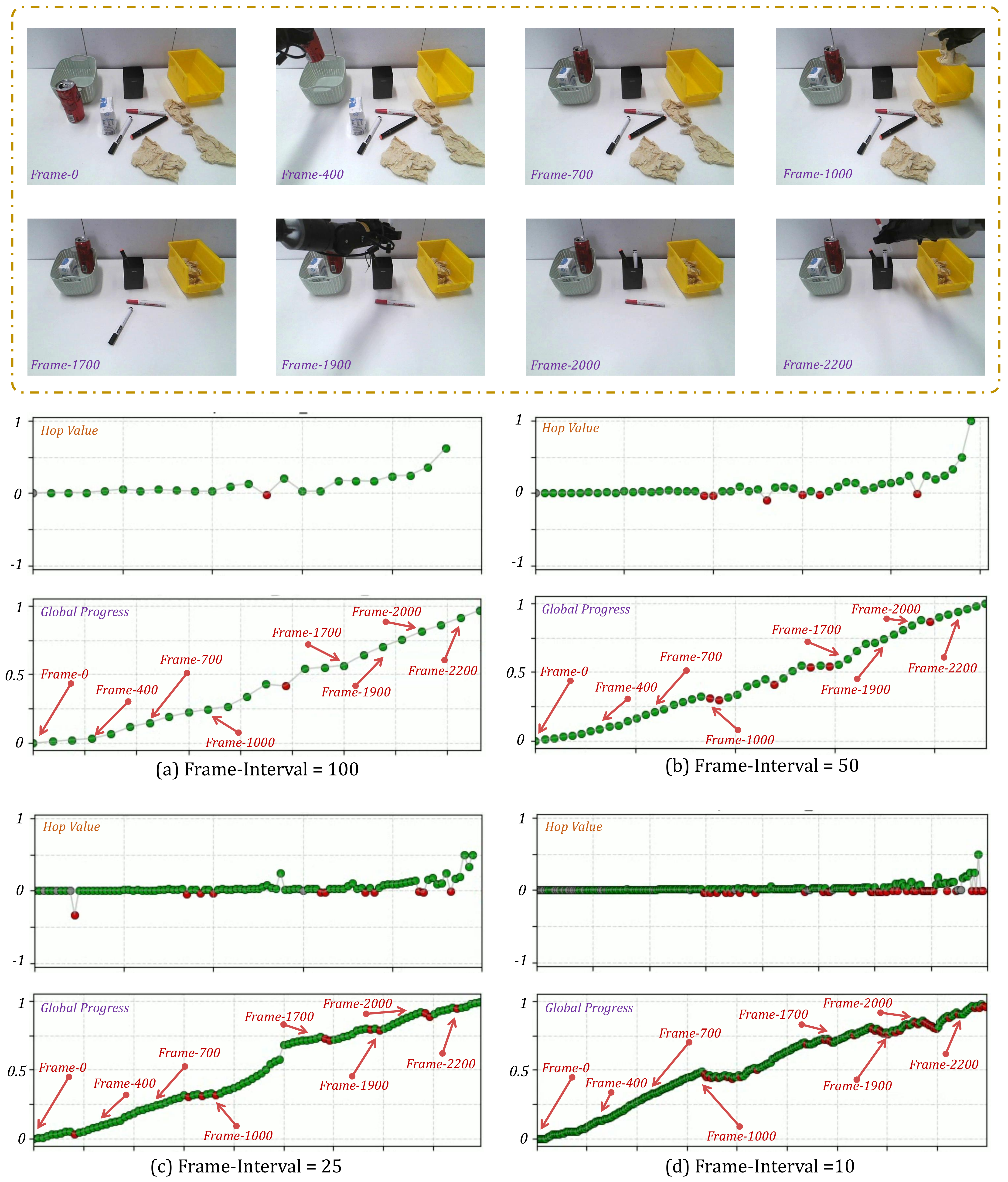}
    \caption{\textbf{Progress Estimation Consistency across Sampling Intervals.} We plot the reconstructed progress curves for the same trajectory using different frame strides (10, 25, 50, and 100 frames). The high overlap between curves demonstrates that our GRM is robust to temporal granularity and does not simply overfit to a specific frame rate.}
    \label{fig:vis_intervals}
\end{figure*}

\textbf{Robustness to Temporal Intervals.}
A robust reward model should remain consistent regardless of the video sampling rate. Figure~\ref{fig:vis_intervals} compares the GRM's progress estimation when inputs are sampled at different intervals ($\Delta t = 10, 25, 50, 100$ frames). Despite the significant variation in visual disparity between adjacent frames, our Hop-based formulation ensures that the reconstructed global progress remains consistent, highlighting the model's ability to decouple physical progress from temporal duration.

\begin{figure*}[t]
    \centering
    \includegraphics[width=0.98\linewidth]{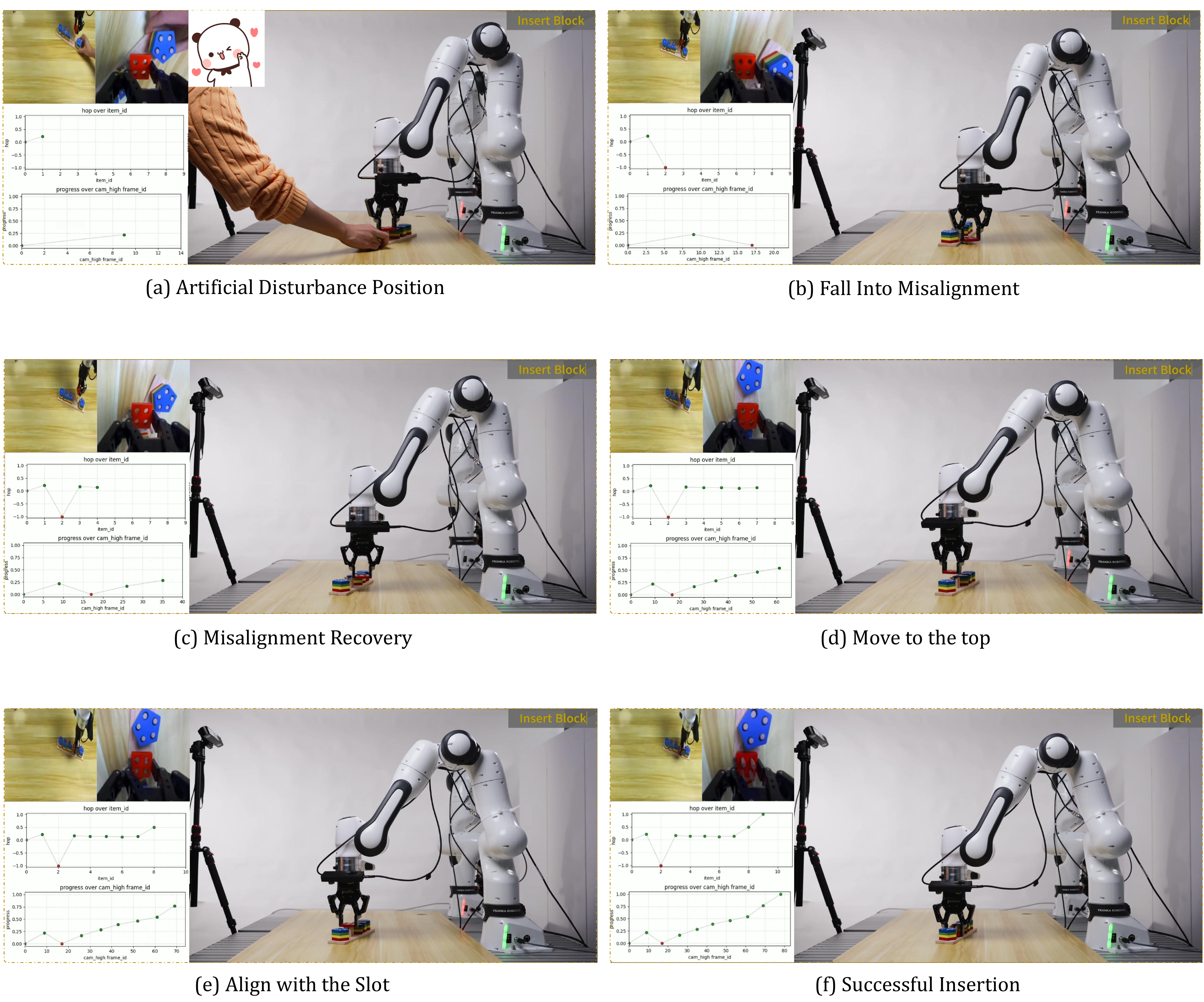}
    \caption{\textbf{Robustness to Artificial Disturbance during Real-World Execution.} We visualize a rollout of the converged policy (success rate $>95\%$) under human interference. Each sub-figure shows the third-person view, the ego-centric view, and the real-time GRM inference (Top: \textit{Hop}, Bottom: \textit{Progress}).
    \textit{\textbf{(a) Artificial Disturbance Position:}} A human hand intervenes and shifts the target board while the robot attempts to approach.
    \textit{\textbf{(b) Fall Into Misalignment:}} The robot misses the new position. Note that the GRM \textit{Progress} curve drops significantly (indicated by the red dot in the bottom inset), reflecting the failure state.
    \textit{\textbf{(c) Misalignment Recovery:}} The policy reacts to the visual feedback and the drop in reward, adjusting the end-effector position.
    \textit{\textbf{(d) Move to the top:}} The robot realigns directly above the target slot.
    \textit{\textbf{(e) Align with the Slot:}} Precise fine-tuning before insertion.
    \textit{\textbf{(f) Successful Insertion:}} The task is completed, with the progress estimation reaching its peak.}
    \label{fig:vis_rl}
\end{figure*}

\textbf{Real-World RL Rollout Visualization.}
Finally, Figure~\ref{fig:vis_rl} visualizes the robustness of the policy learned for the ``Insert Block'' task. The policy used in this rollout was trained for approximately 20 minutes and achieved a success rate of over 95\%. To evaluate the policy's reactivity and the reward model's accuracy, we introduced an artificial disturbance during execution. As shown in the sequence, a human operator manually moves the target slot while the robot is in motion (a). This intervention causes the robot to miss the target and fall into misalignment (b). Crucially, the inset plots show that our Generative Reward Model (GRM) immediately reflects this setback: the estimated \textit{Progress} drops sharply, correctly identifying that the state has regressed from the goal. This accurate negative feedback guides the agent to adjust its trajectory. The robot successfully recovers from the misalignment (c), repositions itself above the target (d), aligns with the slot (e), and completes the insertion (f). This demonstrates that GRM provides dense, semantically meaningful rewards that enable the agent to recover from unexpected external perturbations.

\section{Future Work}
\label{future}

In future research, we plan to expand the capabilities and efficiency of Robo-Dopamine in four primary directions:

\begin{itemize}
    \item \textbf{Efficient GRM Inference:} The current VLM-based reward model, while accurate, incurs high computational latency which can bottleneck online RL training loops. We aim to investigate low-precision quantization techniques (\textit{e.g.,} INT4/INT8 quantization or KV-cache compression) to significantly accelerate GRM inference and reduce memory footprint without compromising reward accuracy for RL phase.

    \item \textbf{Continuous Video Stream Reasoning:} We aim to further evolve the GRM from discrete frame-pair inference to continuous video stream reasoning. By incorporating historical context sequences and integrating temporal modeling architectures (\textit{e.g.,} Video Transformers, Temporal CNNs, or Video-LLM~\cite{xu2025streamingvlm, an2025llava}), the model will gain the ability to capture dynamic motion trends, such as inertia and trajectory, which are essential for high-dynamic tasks. Furthermore, this temporal continuity resolves state ambiguities inherent in static snapshots (\textit{e.g.,} distinguishing between ``grasping'' and ``releasing'' an object, or tracking cumulative progress in cyclic tasks like scrubbing a test tube where multiple repetitions are required), thereby significantly enhancing the robustness of reward assessment.
    
    \item \textbf{Large-Scale Generalization:} We plan to scale the Dopamine-RL framework to a broader range of tasks in both simulation and the real world. Specifically, we aim to validate the framework on highly dynamic tasks (\textit{e.g.,} throwing, catching) and long-horizon mobile manipulation scenarios to test the limits of the policy-invariant reward shaping mechanism.
    
    \item \textbf{Multi-Modal Reward Modeling:} While vision provides rich global context, fine-grained manipulation often requires non-visual cues. We intend to incorporate tactile and auditory modalities into the GRM. Tactile feedback is crucial for contact-rich tasks (\textit{e.g.,} sensing force during insertion), while audio can detect discrete events (\textit{e.g.,} the ``click'' of a latch), enabling more precise reward shaping for delicate operations.
\end{itemize}

\end{document}